\pdfoutput=1
\documentclass[10pt,twocolumn,letterpaper]{article}
\usepackage[letterpaper,textwidth=7in,textheight=9in,columnsep=0.375in,centering]{geometry}
\usepackage[T1]{fontenc}
\usepackage{newtxtext}
\usepackage{helvet}
\usepackage{courier}
\usepackage[hyphens]{url}
\usepackage{graphicx}
\usepackage[round]{natbib}
\usepackage[font=small,labelfont=bf]{caption}
\usepackage{newfloat}
\DeclareFloatingEnvironment[
  name=Algorithm,
  placement=t
]{algorithm}

\usepackage{booktabs}
\usepackage{multirow}
\usepackage{array}
\usepackage{amsmath,amssymb,mathtools}
\usepackage{xspace}
\usepackage{xcolor}
\usepackage{pifont}
\usepackage{tikz}
\usetikzlibrary{arrows.meta,positioning,fit,calc}
\usepackage{pgfplots}
\pgfplotsset{compat=1.18}
\usepackage[hidelinks,pdftitle={FounRef: Robust, Structure-Preserving, and Fast Metric Refinement of Frozen Monocular Foundation Priors with Sparse Anchors},pdfauthor={Dan Halperin, Mirko M\"ahlisch},pdfsubject={},pdfkeywords={},pdfcreator={},pdfproducer={}]{hyperref}
\let\cite\citep

\title{FounRef: Robust, Structure-Preserving, and Fast Metric Refinement of Frozen Monocular Foundation Priors with Sparse Anchors}
\author{Dan Halperin \qquad Mirko M\"ahlisch\\[4pt]
{\normalsize University of the Bundeswehr Munich, Germany}}
\date{}

\begin{document}
\maketitle

\begin{abstract}
Dense metric depth from cameras is essential to real-world 3D applications, yet achieving accuracy, faithful surface geometry, and fast inference simultaneously remains challenging. Monocular foundation models provide rich, transferable geometric priors but lack reliable metric scale, while depth-completion networks recover metric depth at the cost of geometric fidelity, cross-domain robustness, or speed. We present FounRef, a training-free method that aligns a frozen monocular foundation prior with sparse metric anchors to produce dense metric depth. FounRef is modular by design: its depth prior, anchor source, and refinement solver can each be replaced independently. We instantiate FounRef with MoGe-2 and LiDAR anchors. FounRef validates each anchor against the prior’s dense depth prediction, rejecting inconsistencies caused by cross-sensor misalignment that geometry-only filters cannot detect. It then applies global and local metric corrections through a structure-preserving solver, retaining the prior’s fine-grained geometry. FounRef requires no task-specific training and operates out of the box across unfamiliar cameras and scenes. On out-of-domain data, it delivers up to 24\% lower depth error, 92\% lower surface-normal noise, and almost 15× faster inference than DMD3C, a state-of-the-art depth-completion network. By decoupling metric alignment from geometry prediction, FounRef provides an accurate, geometrically faithful, and efficient approach to dense metric depth that can directly benefit from future advances in foundation models and metric sensors.
\end{abstract}
\section{Introduction}
\label{sec:intro}

\begin{figure}[t]
  \centering
  \includegraphics[width=\linewidth]{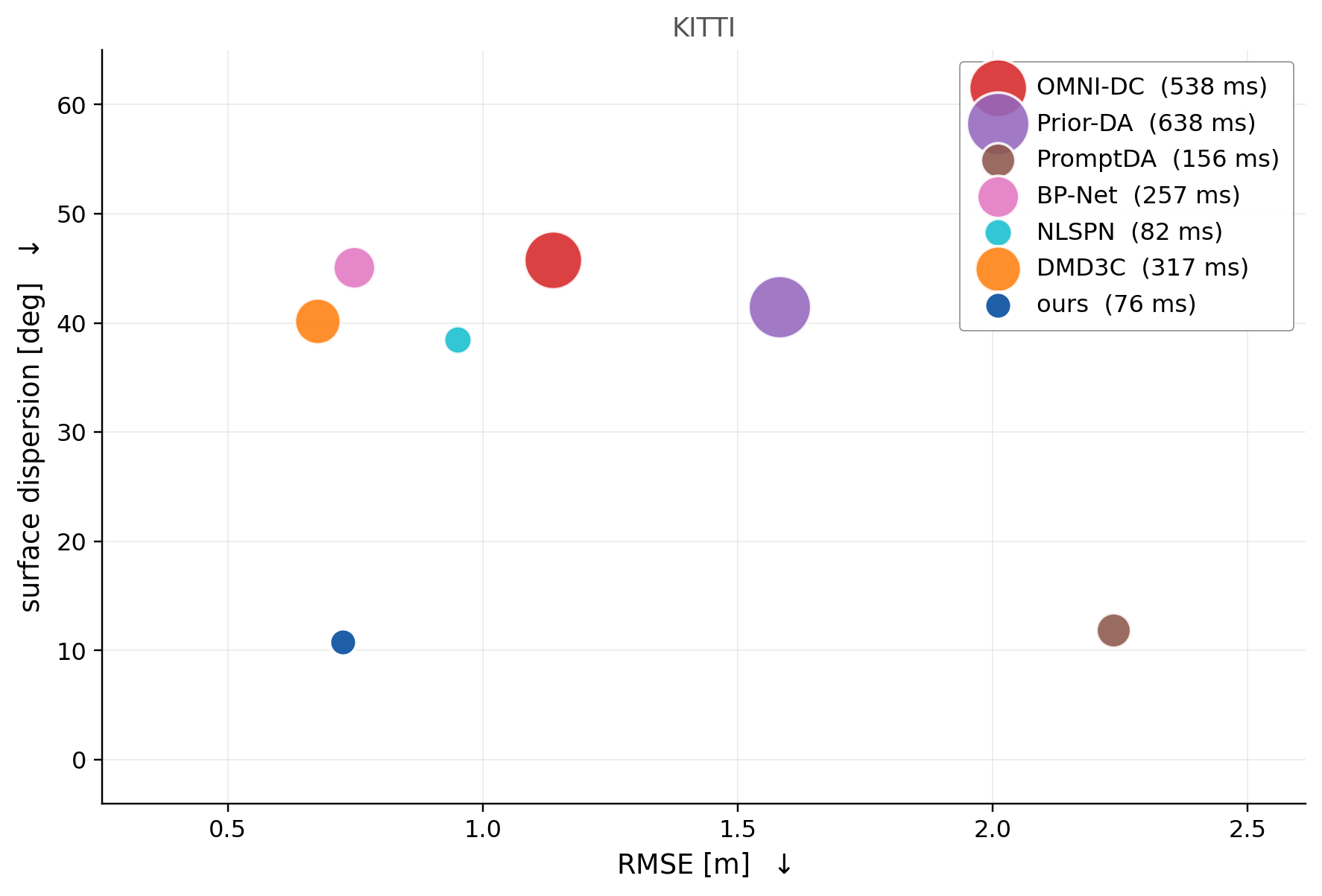}
  \caption{\textbf{Accuracy is not the only axis.} Metric error against surface dispersion on KITTI
  ($n{=}500$ frames, identical anchors for every method); marker area grows with runtime. A single configuration change moves our point to $0.663$\,m at $11.2^\circ$
  without affecting runtime}
  \label{fig:teaser}
\end{figure}

Estimating metric depth from a single image is fundamentally ill posed. An image provides strong cues about the relative structure of a scene, but it does not provide enough geometric information to determine metric scale reliably across arbitrary scenes and camera settings~\cite{zerodepth}. Nevertheless, dense metric depth is essential for real-world applications such as autonomous systems, mapping, and the construction of 3D datasets.

Recent monocular foundation models produce increasingly detailed and transferable depth
predictions, such as MoGe-2~\cite{moge2}, UniDepth~\cite{unidepth}, DepthPro~\cite{depthpro} and Metric3D-V2~\cite{metric3dv2}. They often recover sharp boundaries and plausible surface geometry on unfamiliar scenes. However, their metric calibration remains imperfect, especially under domain shift and most notably at long ranges. In many practical settings, a sensor rig provides only an RGB image, camera intrinsics, and sparse metric anchors, such as a LiDAR scan projected into the image. The challenge is therefore not only to obtain metric depth, but also to preserve object boundaries and smooth physical surfaces while remaining efficient enough to process large image collections.

Depth-completion networks address this problem by learning to combine RGB images and sparse depth measurements, from early sparse-to-dense regression~\cite{ma2018sparse2dense} to recent models such as DMD3C~\cite{dmd3c} and OMNI-DC~\cite{omnidc}. Their accuracy can be high, but it is tied to the training procedure, the available training data, and the sparse-depth patterns seen during training; adapting them to a new sensor rig or data distribution may require retraining or fine-tuning. Moreover, as illustrated in Figure~\ref{fig:teaser}, optimizing metric error alone does not guarantee coherent surfaces: a method can improve depth metrics while introducing artifacts plainly visible in normal maps and at object boundaries (Figure~\ref{fig:exp4_bakeoff_qual}). 

To avoid this trade-off, FounRef uses a frozen monocular foundation model as a geometric prior, sparse metric anchors as direct observations, and a lightweight per-image refinement procedure. It globally calibrates the prior, validates anchors against a locally refined reference surface, and propagates the retained corrections along surfaces while preserving depth discontinuities. Our algorithm refines the prior rather than replacing it, so its output remains limited by the prior where anchors are sparse; better priors lead to better refined depth maps. Retraining for a new sensor setup is expensive and can cause a model to fit sparse, noisy anchor points rather than preserve the surface between them. Instead, FounRef exposes metric evidence per image without retraining, while retaining coherent geometry for downstream tasks.
\section{Related Work}
\label{sec:related}

\paragraph{Depth datasets and evaluation.}
The KITTI depth-completion benchmark~\cite{uhrig2017sparsity}, built on the
KITTI dataset~\cite{kitti}, and NYU Depth V2~\cite{nyuv2} provide refined
dense targets for depth-completion evaluation. In contrast,
nuScenes~\cite{nuscenes}, the Waymo Open Dataset~\cite{waymo}, and
GOOSE~\cite{goose} retain the sparse and imperfect projected measurements
encountered in newly collected data. Together, they test transfer across
sensor configurations, scene types, and target quality.

\paragraph{Monocular depth foundation models.}
Modern monocular models, such as  Depth Anything V2~\citep{depthanythingv2}, Depth Anything V3 \cite{lin2025depthanything3} Metric3D v2~\citep{metric3dv2}, UniDepth~\citep{unidepth}, Depth Pro~\citep{depthpro}, and MoGe-2~\citep{moge2}, provide detailed dense geometry. However, monocular models are still biased metrically, particularly under domain shift and at long range. Thus, FounRef uses this geometry as a frozen prior and grounds it with direct metric LiDAR anchors.

\paragraph{Learned depth completion.}
Depth-completion methods predict dense metric depth from an RGB image and sparse depth observations~\citep{ma2018sparse2dense}. They learn image-guided depth propagation, for example through non-local affinities in NLSPN~\citep{nlspn}, transformer context in CompletionFormer~\citep{completionformer}, or foundation-model distillation in DMD$^3$C~\citep{dmd3c}. Other foundation-model conditioning methods include PromptDA \cite{lin2025promptda} and Prior Depth Anything \cite{wang2025priorda}, while BP-Net \cite{tang2024bpnet} is a propagation-based completion baseline. These learned, scene-dependent priors require offline training and depend on the training distribution, sensor configuration, and input sparsity. On the other hand, OMNI-DC~\citep{omnidc} targets zero-shot transfer across datasets and sparse-depth patterns. In contrast, FounRef is training-free and adapts a frozen monocular prior to each image using only sparse metric anchors.

\paragraph{Optimization-based depth refinement.}
Explicit refinement methods propagate sparse depth constraints through a regularized correction field. TGV$^2$~\citep{tgv} regularizes both the correction gradient and its spatial change, favoring locally affine corrections, while screened Poisson refinement~\citep{screenedpoisson} balances anchor fitting with spatial smoothness. The fast bilateral solver~\citep{fastbilateralsolver} instead solves an edge-aware correction on a low-dimensional bilateral grid, allowing efficient propagation within similar regions while limiting propagation across boundaries.

\paragraph{Anchor outlier handling.}
Projected LiDAR anchors can be corrupted by occlusion, calibration error, moving objects, temporal registration, and projection artifacts. KITTI provides a refined dense target for offline evaluation, obtained by multimodal careful voting, but such a target is unavailable when processing data from a new sensor rig. RePLAy~\citep{replay2024} uses a parameter-free epipolar-geometry test to remove baseline-induced projective occlusions. FounRef is complementary: it tests each anchor against a locally refined dense reference surface, allowing it to reject disagreements caused by a broader range of cross-modal and temporal inconsistencies.
\section{Method}
\label{sec:method}

\begin{algorithm}[t]
\caption{FounRef pipeline}
\label{alg:founref}
\begin{enumerate}

    \item Predict the frozen monocular prior and its validity mask:
    $(D_{\mathrm{prior}}, M) = \mathrm{Prior}(I)$.

    \item Cap the anchors at $D_{\max}=50m$ and split them into fixed fitting and held-out sets:
    $(\mathcal{A}_{\mathrm{fit}}, \mathcal{A}_{\mathrm{test}})
    = \mathrm{Split}(\mathrm{Cap}(\mathcal{A}, D_{\max}), 80/20)$.

    \item Globally calibrate the prior using the fitting anchors:
    $D_{\mathrm{cal}} =
    \mathrm{Calibrate}(D_{\mathrm{prior}}, \mathcal{A}_{\mathrm{fit}})$.

    \item If validation is enabled, construct the locally corrected reference surface:
    $D_1 =
    \mathrm{LightSolve}(D_{\mathrm{cal}}, \mathcal{A}_{\mathrm{fit}})$.

    \item Filter fitting and held-out anchors against this reference:
    $\mathcal{A}^{+} =
    \{(\mathbf{u}_i,z_i)\in\mathcal{A} :
    |\log z_i-\log D_1(\mathbf{u}_i)|\leq\tau\}$.

    \item Run the full solver using only retained fitting anchors:
    $D_2 =
    \mathrm{FullSolve}(D_{\mathrm{cal}}, \mathcal{A}_{\mathrm{fit}}^{+})$.

\end{enumerate}
\end{algorithm}

\begin{figure*}[t]
  \centering
  \setlength{\tabcolsep}{1pt}
  \renewcommand{\arraystretch}{0}
  \begin{tabular}{@{}ccccc@{}}
    \includegraphics[width=0.19\linewidth]{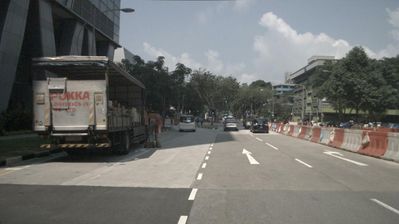} &
    \includegraphics[width=0.19\linewidth]{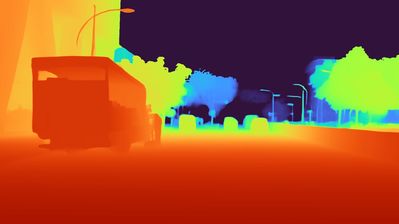} &
    \includegraphics[width=0.19\linewidth]{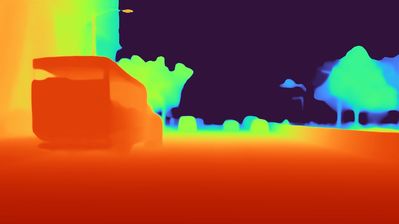} &
    \includegraphics[width=0.19\linewidth]{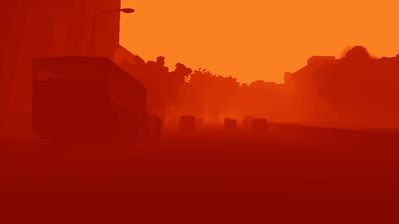} &
    \includegraphics[width=0.19\linewidth]{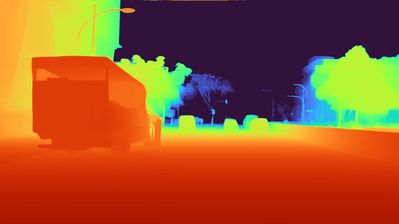} \\[3pt]
    \scriptsize RGB &
    \scriptsize MoGe-2 (raw) &
    \scriptsize Metric3D-V2 (raw) &
    \scriptsize DA-V2 (raw) &
    \scriptsize DepthPro (raw) \\[9pt]
    \includegraphics[width=0.19\linewidth]{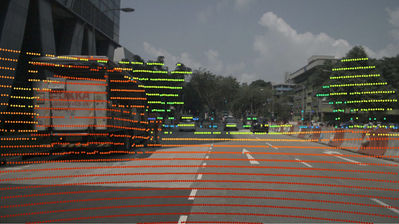} &
    \includegraphics[width=0.19\linewidth]{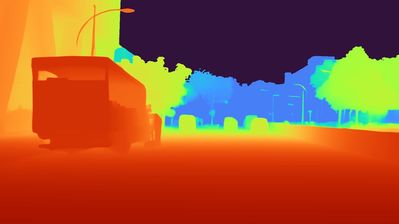} &
    \includegraphics[width=0.19\linewidth]{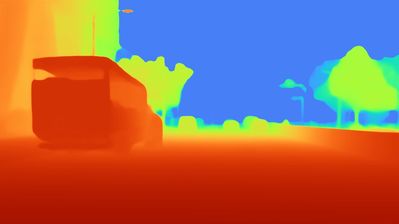} &
    \includegraphics[width=0.19\linewidth]{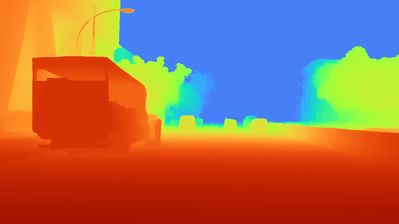} &
    \includegraphics[width=0.19\linewidth]{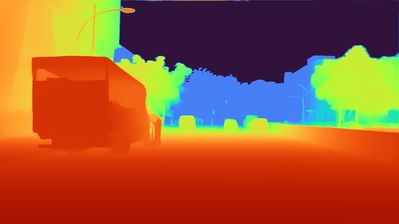} \\[3pt]
    \scriptsize sparse anchors &
    \scriptsize MoGe-2 (calibrated) &
    \scriptsize Metric3D-V2 (calib.) &
    \scriptsize DA-V2 (calibrated) &
    \scriptsize DepthPro (calib.) \\
  \end{tabular}
  \caption{Global calibration of four priors on the same frame. Top: raw predictions use incompatible depth scales; DA-V2 is relative. Bottom: after fitting the same sparse anchors, all priors share a metric scale while preserving their edges and relative layout. All depth panels use the $0$--$80\mathrm{m}$ range. Pixels a prior marks invalid (sky) are drawn at the far end (MoGe-2 provides an invalid map, while depth pro predicts them at $10,000m$).}
  \label{fig:exp1-prior-calibration}
\end{figure*}

FounRef converts a frozen monocular depth prior into a dense metric depth map using sparse metric anchors. Given an input RGB image $I$, a frozen foundation model predicts a dense prior depth map $D_{\mathrm{prior}}$ and, when available, a validity mask $M$. The refinement stage receives the prior and a set of projected metric anchors
$
\mathcal{A} = \{(\mathbf{u}_i, z_i)\}_{i=1}^{N},
$
where $\mathbf{u}_i$ is an image location and $z_i$ is its measured metric depth. The output is a dense metric depth map $D$. Pixels marked invalid by the prior are excluded from fitting and assigned the maximum depth cap $D_{\max}$ in the final output. The algorithm is defined in Algorithm~\ref{alg:founref}.

\paragraph{Global monotone calibration.}
\label{sec:calibration}

The prior predicts depth on the wrong scale. We correct this in log space,
where a scale factor becomes a shift. Writing $t_i=\log D_{\mathrm{prior}}(\mathbf{u}_i)$
and $y_i=\log z_i$ for each fitting anchor, we fit a straight line
$f(t)=\alpha t+\beta$:
\begin{equation}
(\hat{\alpha},\hat{\beta})=
\underset{\alpha,\beta}{\operatorname*{arg\,min}}
\sum_i \rho_{\mathrm{Huber}}\!\left(y_i-f(t_i)\right)
\label{eq:backbone-fit}
\end{equation}
where $\rho_{\mathrm{Huber}}$ is the Huber loss, solved with iteratively reweighted least squares ~\cite{huber1964robust}.

One line cannot fit every distance equally well. We therefore split the anchors
into $K$ bins by depth and record the average error the line makes in each bin.
Interpolating these $K$ values gives a small per-distance correction $c$, and
the calibrated prior is
\begin{equation}
\log D_{\mathrm{cal}}(\mathbf{x})
=
f\!\left(\log D_{\mathrm{prior}}(\mathbf{x})\right)
+
c\!\left(\log D_{\mathrm{prior}}(\mathbf{x})\right)
\label{eq:monotone-calibration}
\end{equation}
We use $K=24$, chosen once on validation as the smallest value that saturated calibration accuracy; the mapping is monotone, sparse bins are damped, and extrapolation is disabled. With too few anchors we use $f$ alone; with none, we keep the prior.

\paragraph{Anchor validation.}
\label{sec:anchor-validation}
To filter cross-modal noise, we validate anchors before the final refinement. Validation requires a metric surface that is locally more accurate than global calibration alone. We obtain this surface by running a lightweight local solver on the fitting anchors from the current sweep only, producing $D_1$. For an anchor $i$ projected at $\mathbf{u}_i$, we retain the anchor when
\begin{equation}
\left|
\log z_i - \log D_1(\mathbf{u}_i)
\right|
\leq \tau,
\label{eq:anchor-validation}
\end{equation}
and discard it otherwise, and where $\tau$ is a dataset-specific threshold in log space. Because the test operates in log depth, it measures relative disagreement rather than an absolute error that increases with range. We select that threshold $\tau$ from a validation operating curve. Cleaner datasets, such as KITTI~\cite{kitti}, permit a more relaxed threshold, whereas noisier datasets, such as nuScenes~\cite{nuscenes} and GOOSE~\cite{goose}, require stricter filtering.

The aim is to discard only a small fraction of unreliable anchors while preserving a dense metric point cloud. This is particularly important under temporal accumulation: additional sweeps improve coverage only when their outliers are removed. Since FounRef is training-free, it cannot learn to suppress these errors from data; explicit anchor validation is therefore necessary. 

\paragraph{Structure-preserving local solver.}
\label{sec:solver}

We refine the calibrated prior with a single additive correction field in log depth:
\begin{equation}
\log D(\mathbf{x})
=
\log D_{\mathrm{cal}}(\mathbf{x})
+
b(\mathbf{x})
\label{eq:log-shift}
\end{equation}
where $b$ is a smooth correction field. Thus, the solver adjusts the prior's surfaces rather than redrawing their structure. For anchor $i$, the desired correction is
\begin{equation}
t_i
=
\log z_i
-
\log D_{\mathrm{cal}}(\mathbf{u}_i).
\end{equation}
We estimate $b$ by minimizing
\begin{equation}
E(b)
=
\sum_i
w_i
\bigl(
b(\mathbf{u}_i) - t_i
\bigr)^2
+
\lambda b^\top L b\, ,
\label{eq:solver-energy}
\end{equation}
Here, $b(u_i)$ is the correction interpolated from the bilateral grid at
anchor $u_i$, $w_i$ is its confidence, and $\lambda$ sets the strength of
the bilateral smoothness prior. The bilateral Laplacian ~\cite{fastbilateralsolver}
penalizes differences between corrections at vertices that are nearby in image position and similar in prior depth. Thus, anchors share a correction along a predicted surface,
while predicted depth discontinuities limit propagation across surfaces.

We construct $L$ on a bilateral grid over image position and log prior depth, and five bistochastic normalization steps~\cite{fastbilateralsolver}. Two pixels are coupled only when they are close in the image and lie at similar prior depth. A correction can therefore spread smoothly along a surface while stopping at its depth boundaries: for example, an anchor on a nearby pole corrects the pole without pulling the background towards it.

The resulting sparse positive-definite system is solved on grid vertices with Jacobi-preconditioned conjugate gradients. We solve at half resolution, aggregating anchors in each $2\times2$ block as a weighted mean in log-depth so that returns from different surfaces never average into a depth that belongs to
neither. Only the correction is upsampled and applied to the full-resolution prior, which keeps the prior's detail intact. The weights $w_i$ combine agreement with a robust local consensus and a depth-dependent confidence term, so the solver trusts the prior near the camera and the anchors where the prior starts to drift.

We use a pure shift field rather than a local scale-and-shift field. We also exclude image intensity from the bilateral grid: although it can improve metric accuracy, it can imprint image texture into the recovered geometry. 
\section{Experiments}
\label{sec:setup}
\begin{figure*}[t]
  \centering
  \setlength{\tabcolsep}{1pt}
  \renewcommand{\arraystretch}{0}
  \begin{tabular}{@{}ccccccc@{}}
    \includegraphics[width=0.139\linewidth]{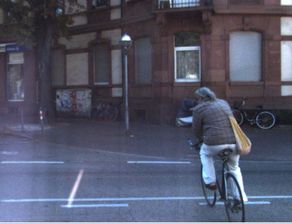} &
    \includegraphics[width=0.139\linewidth]{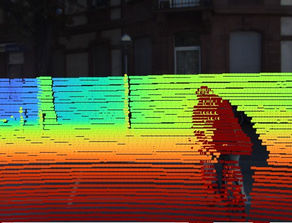} &
    \includegraphics[width=0.139\linewidth]{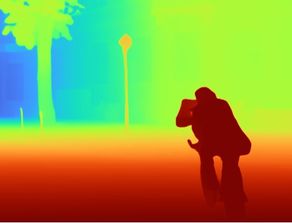} &
    \includegraphics[width=0.139\linewidth]{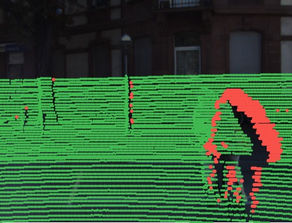} &
    \includegraphics[width=0.139\linewidth]{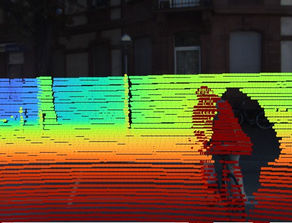} &
    \includegraphics[width=0.139\linewidth]{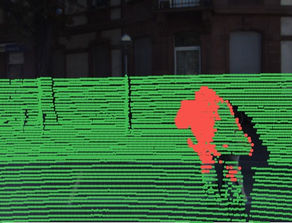} &
    \includegraphics[width=0.139\linewidth]{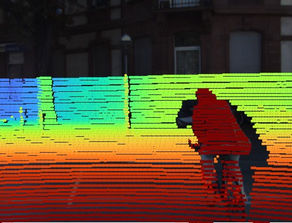} \\[4pt]
    \scriptsize RGB &
    \scriptsize raw anchors &
    \scriptsize prior $D_1$ (ref.) &
    \scriptsize RePLAy mask &
    \scriptsize RePLAy filtered &
    \scriptsize ours mask &
    \scriptsize ours filtered \\
  \end{tabular}
  \caption{Anchor filtering on a cluttered KITTI crop. RePLAy retains returns through windshields, mirrors, and moving-vehicle silhouettes, whereas FounRef rejects them because they disagree with the locally refined smooth surface $D_1$.}
  \label{fig:exp2_filter_qual}
\end{figure*}
\begin{figure*}[t]
  \centering
  \setlength{\tabcolsep}{1pt}
  \renewcommand{\arraystretch}{0}
  \begin{tabular}{@{}ccccc@{}}
    \includegraphics[width=0.196\linewidth]{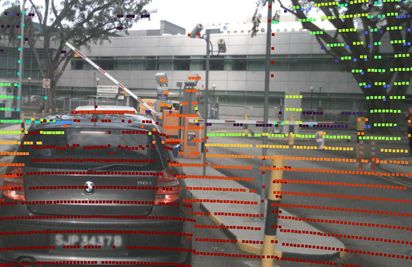} &
    \includegraphics[width=0.196\linewidth]{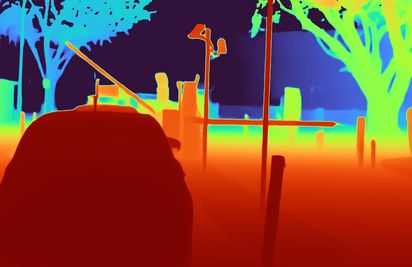} &
    \includegraphics[width=0.196\linewidth]{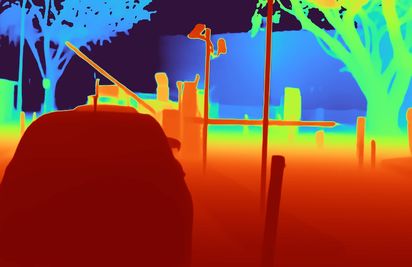} &
    \includegraphics[width=0.196\linewidth]{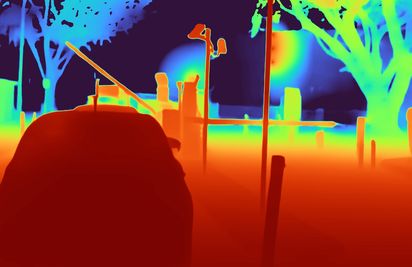} &
    \includegraphics[width=0.196\linewidth]{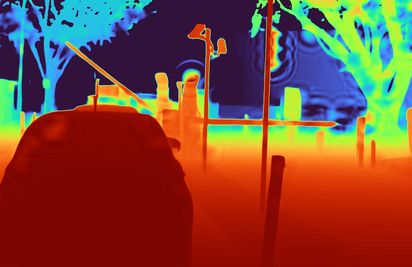} \\[4pt]
    \scriptsize RGB &
    \scriptsize calibrated prior &
    \scriptsize ours (bilateral) &
    \scriptsize Poisson &
    \scriptsize TGV \\
  \end{tabular}
  \caption{Solver comparison on a sparse-anchor nuScenes crop. All methods start from the same calibrated prior and anchors. Pixel-wise Poisson and TGV produce anchor-centred surface bumps, whereas FounRef shares corrections on a coarse bilateral grid. Depth panels share the $[1,50]\,\mathrm{m}$ range.}
  \label{fig:exp3_solvers_qual}
\end{figure*}
\begin{figure*}[!t]
  \centering
  \setlength{\tabcolsep}{1pt}
  \renewcommand{\arraystretch}{0}
  \begin{tabular}{@{}ccccc@{}}
    \includegraphics[width=0.192\linewidth]{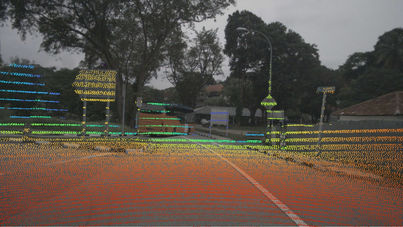} &
    \includegraphics[width=0.192\linewidth]{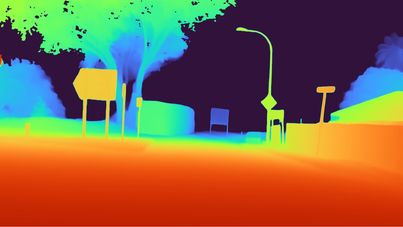} &
    \includegraphics[width=0.192\linewidth]{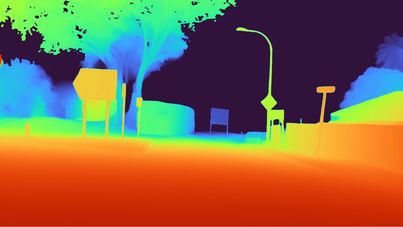} &
    \includegraphics[width=0.192\linewidth]{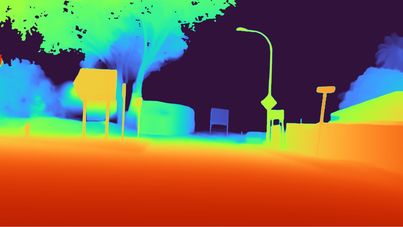} &
    \includegraphics[width=0.192\linewidth]{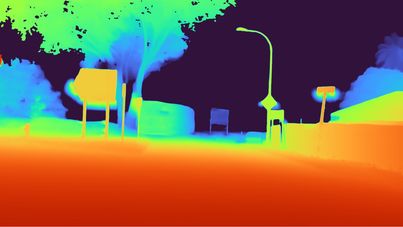} \\[3pt]
    \scriptsize RGB + anchors &
    \scriptsize \textbf{shipped} $\sigma_s16,\lambda10$ &
    \scriptsize $\sigma_s8,\lambda5$ &
    \scriptsize $\sigma_s4,\lambda2.5$ &
    \scriptsize $\sigma_s2,\lambda1$ \\
    & \scriptsize 0.687 / 3.5$^\circ$ & \scriptsize 0.577 / 3.6$^\circ$
    & \scriptsize 0.489 / 3.6$^\circ$ & \scriptsize 0.418 / 8.0$^\circ$ \\
  \end{tabular}
  \caption{One solver, four operating points on nuScenes ($0$--$50\,\mathrm{m}$). $\sigma_s$ is the bilateral grid’s spatial bandwidth in pixels; reducing it localizes corrections, improving RMSE but increasing normal dispersion.}
  \label{fig:tradeoff}
\end{figure*}

\begin{table*}[t]
  \centering
  \setlength{\tabcolsep}{2pt}
  \caption{Depth completion across five validation sets ($n=500$) using
  identical anchors for every method and no dataset-specific recalibration.
  KITTI and NYU use dense ground truth; the other datasets use a held-out
  $20\%$ of the cloud. RMSE and far (RMSE over target depths
  $30 < z \leq 50\,\mathrm{m}$) are in metres, \emph{surf.} is normal
  dispersion ($^\circ$), and \emph{ms} is mean latency on the Waymo
  validation set, our worst-case setting, on an A100 (most models cannot
  run at fp16 precision).}
  \label{tab:bakeoff}
  \begin{tabular}{lcccccccccccccccc}
    \toprule
    & \multicolumn{3}{c}{KITTI} & \multicolumn{3}{c}{nuScenes}
    & \multicolumn{3}{c}{Waymo} & \multicolumn{3}{c}{GOOSE}
    & \multicolumn{3}{c}{NYU} & \\
    \cmidrule(lr){2-4}\cmidrule(lr){5-7}\cmidrule(lr){8-10}
    \cmidrule(lr){11-13}\cmidrule(lr){14-16}
    & RMSE$\downarrow$ & far$\downarrow$ & surf.$\downarrow$
    & RMSE$\downarrow$ & far$\downarrow$ & surf.$\downarrow$
    & RMSE$\downarrow$ & far$\downarrow$ & surf.$\downarrow$
    & RMSE$\downarrow$ & far$\downarrow$ & surf.$\downarrow$
    & RMSE$\downarrow$ & far$\downarrow$ & surf.$\downarrow$
    & ms$\downarrow$ \\
    \midrule
    \textbf{Ours}
      & 0.726 & 2.08 & \textbf{10.8}
      & 0.687 & 2.00 & \textbf{3.5}
      & 0.745 & 1.43 & \textbf{1.7}
      & 1.687 & 4.42 & \textbf{8.8}
      & 0.174 & 0.40 & \textbf{6.6}
      & \textbf{78} \\
    \textbf{Ours} ($\sigma_s$4)
      & \textbf{0.663} & \textbf{1.83} & 11.2
      & 0.489 & 1.27 & 3.6
      & \textbf{0.587} & \textbf{1.10} & 2.2
      & 1.073 & 3.24 & 8.9
      & 0.164 & 0.37 & 6.7
      & \textbf{78} \\
    \midrule
    OMNI-DC
      & 1.137 & 3.61 & 45.8
      & \textbf{0.353} & \textbf{0.87} & 34.6
      & 0.590 & 1.12 & 41.7
      & 0.702 & 2.01 & 33.1
      & \textbf{0.099} & \textbf{0.20} & 10.2
      & 1011 \\
    Prior-DA
      & 1.583 & 4.83 & 41.4
      & 0.661 & 2.19 & 16.6
      & 0.833 & 1.69 & 13.2
      & 0.742 & 2.03 & 15.8
      & 0.135 & 0.25 & 13.0
      & 1621 \\
    PromptDA
      & 2.237 & 7.32 & 11.9
      & 2.058 & 7.41 & 8.8
      & 2.770 & 6.48 & 5.8
      & 1.545 & 4.91 & 11.0
      & 0.218 & 0.40 & 11.1
      & 580 \\
    BP-Net
      & 0.747 & 2.11 & 45.1
      & 0.602 & 1.66 & 35.0
      & 0.932 & 1.53 & 49.2
      & \textbf{0.693} & 1.85 & 30.6
      & 1.055 & 0.79 & 23.3
      & 1081 \\
    NLSPN
      & 0.951 & 2.47 & 38.5
      & 0.834 & 2.11 & 24.3
      & 0.756 & 1.39 & 27.3
      & 0.764 & \textbf{1.82} & 27.6
      & 2.720 & 0.76 & 21.8
      & 347 \\
    DMD3C
      & 0.677 & 1.92 & 40.1
      & 0.523 & 1.53 & 18.7
      & 0.774 & 1.56 & 25.6
      & 0.805 & 1.92 & 23.4
      & 1.205 & 2.10 & 20.6
      & 1219 \\
    \bottomrule
  \end{tabular}
\end{table*}
\begin{table}[t]
  \centering\small
  \setlength{\tabcolsep}{4pt}
  \renewcommand{\arraystretch}{1.1}
  \caption{Solver comparison on KITTI val against accumulated dense depth.
  All solvers receive identical prior, calibration, anchors, and weights;
  only the regularizer differs. \emph{RMSE}: error over all valid pixels;
  \emph{far}: RMSE in the $30$--$50$\,m band. \emph{surf.}: median angle
  between the output normals and calibrated prior normals in smooth regions.
  \emph{imp.}: correction at anchor pixels relative to a few pixels away,
  so $1.0$ leaves no trace of the anchor pattern.}
  \label{tab:exp3_solvers}
  \begin{tabular}{lccccc}
    \toprule
    solver & RMSE\,$\downarrow$ & far\,$\downarrow$
           & surf.\,$\downarrow$ & imp.\,$\rightarrow\!1$ & ms\,$\downarrow$ \\
    \midrule
    \textbf{Ours}
      & 1.050 & 2.583 & \textbf{1.77$^\circ$} & \textbf{0.38} & \textbf{14} \\
    Screened Poisson
      & \textbf{0.880} & \textbf{1.904} & 4.75$^\circ$ & 5.25 & 42 \\
    TGV$^2$-L2
      & 0.934 & 1.949 & 6.32$^\circ$ & --- & 300 \\
    \bottomrule
  \end{tabular}
\end{table}

We run all experiments on a single NVIDIA A100 GPU using PyTorch~$2.10$ and CUDA~$12.8$. Each competing network remains resident in a separate process throughout evaluation on a dataset, running in fp16 precision when possible.

\paragraph{Data.}
We use the five datasets described above. Our deployment setting assumes an RGB image, camera intrinsics when required by the monocular prior, and sparse direct metric-depth anchors, optionally augmented with temporally accumulated anchors when context is available.

Since monocular priors become substantially  less reliable at longer ranges (Table~\ref{tab:exp1_priors}), we cap input anchors at $50\,\mathrm{m}$, where monocular priors remain sufficiently reliable for metric alignment. For full context, we additionally report far-range error beyond this cap. This range retains the challenging medium while avoiding a comparison dominated by unreliable far-range estimates.

When refined dense ground truth is available, namely on KITTI and NYU Depth V2, we report dense-depth metrics for direct comparison with prior work. In addition, we report error on a filtered held-out subset of the raw metric measurements. For every image, we split these measurements deterministically, using a fixed seed of $42$, into $80\%$ fitting anchors and $20\%$ held-out anchors. This sensor-grounded metric does not replace official benchmark metrics; rather, it measures agreement with independent observations that were not used during refinement. It is the primary evaluation protocol for nuScenes, Waymo, and GOOSE, where refined dense ground truth is unavailable. For these datasets, we construct denser metric targets by accumulating measurements from neighboring frames, up to approximately ten frames. Accumulation substantially improves coverage, but also introduces artifacts from moving objects, ego-motion error, occlusions, and sensor noise. Therefore,we filter the accumulated points using the structural consistency of an independent Metric3D-V2 prior, which exhibits strong metric and structural performance in our prior-selection experiment (Table~\ref{tab:exp1_priors}, Figure~\ref{fig:exp1-prior-calibration}). This avoids biasing the evaluation target toward FounRef, which uses MoGe-2 as its refinement prior, while applying the same surface-consistency criterion used to identify outlier anchors. Points that disagree with the estimated visible scene geometry are removed before evaluation, producing cleaner and more spatially coherent targets than unfiltered temporal accumulation. 

Finally, we evaluate 500 images from each dataset's validation set, providing a consistent sample size across domains while keeping all comparisons computationally tractable.

\paragraph{Prior selection.}
The monocular foundation model is the backbone of FounRef: it provides the initial depth estimate, fine geometric structure, and a substantial fraction of the total runtime. We therefore compare several frozen foundation models (Table~\ref{tab:exp1_priors}, Figure~\ref{fig:exp1-prior-calibration}). None is fine-tuned or retrained.

All priors are evaluated on the same KITTI images with identical anchors, global calibration, and local refinement. 

We report three axes. Accuracy on the dense target: RMSE, AbsRel, and RMSE beyond $50$\,m, where priors differ most. Structure, as a pair of spectral measures: \emph{incoherence}, the high-frequency energy on flat surfaces, where any such content is noise, and \emph{detail}, the same quantity on finely textured regions, where it is real scene structure. They have to be read together: a washed-out prediction scores well on incoherence alone, so only a prior that is low on the first and high on the second is both clean and sharp. And runtime, measured per frame.

\paragraph{Solver selection.} The local solver determines how sparse metric corrections propagate across the depth map. We compare the fast bilateral solver, screened edge-aware Poisson refinement and TGV$^2$-L2 refinement. To isolate the solver, the prior, its global calibration, and the per-anchor confidence weights are computed once and handed byte-identically to every method; only the regularizer differs, and all four solve the same unknown, the log-residual $r = \log(D / D_{\mathrm{cal}})$. All methods run on the same 100 KITTI frames (seed 42), scored against KITTI’s accumulated dense depth maps (Table~\ref{tab:exp3_solvers}; qualitative results and the bandwidth trade-off appear in Figures ~\ref{fig:exp3_solvers_qual} and ~\ref{fig:tradeoff}.

\paragraph{Anchor filtering.}
Sparse anchors are not uniformly reliable. Even a single sweep can contain outliers caused by sensor misalignment, moving objects, or occlusions. These errors become more frequent when anchors are accumulated over time. We therefore test every anchor against a reference surface before the final refinement.

The reference surface is produced from the monocular prior in two stages. We first apply global calibration and then refine the calibrated prior with the lightweight local solver, producing the reference depth map $D_1$. An anchor is removed when its log-depth differs from $D_1$ by more than $\tau$. We use $\tau = 0.45$ on KITTI; for the noisier nuScenes data, we use the stricter threshold $\tau = 0.2$, selected from a validation operating curve (See Appendix).

We compare against RePLAy as a representative geometry-based point-cloud filter. Because RePLAy targets projective occlusions using LiDAR geometry, it does not address the broader cross-sensor and temporal inconsistencies captured by our surface-consistency test.

We evaluate the filter on KITTI. First, we report the fraction of capped anchors removed by the filter. Next, using KITTI ground truth, we compare the relative depth error of retained and removed anchors across depth ranges. This reveals whether the filter removes primarily incorrect anchors rather than simply discarding many measurements. Finally, we run the full refinement with and without filtering and report the resulting depth accuracy and runtime. We repeat this evaluation for single-sweep and temporally accumulated anchors because the two settings contain different types and levels of noise (Table~\ref{tab:exp2_filter}, Figure~\ref{fig:exp2_filter_qual}).

\paragraph{Depth-completion benchmark.}
We compare FounRef with supervised sparse-to-dense and zero-shot depth-completion methods.  Each baseline is evaluated using its publicly released checkpoint, without dataset-specific recalibration or fine-tuning, and receives the same sparse input as FounRef.

The benchmark spans five datasets and two complementary settings. KITTI and NYU Depth V2 provide standard benchmarks with refined dense ground truth, including domains on which some baselines were trained. nuScenes, Waymo, and GOOSE assess transfer to previously unseen driving and unstructured outdoor data, where raw sensor measurements are sparse and imperfect. Together, these settings test whether FounRef generalizes across sensor configurations, scene types, and input distributions rather than only within the training domain of a learned completion model.

Importantly, we evaluate both metric accuracy and surface structure. This distinction reveals not only where methods transfer successfully across distributions, but also whether a gain in point-wise accuracy is achieved by overfitting sparse measurements and damaging the depth surface between them. A solver optimized solely to fit anchors could obtain lower point-wise error while introducing local artifacts; our evaluation therefore measures whether metric correction is achieved without sacrificing coherent scene geometry.

\begin{figure*}[t]
  \centering
  \setlength{\tabcolsep}{1pt}
  \renewcommand{\arraystretch}{0}
  \begin{tabular}{@{}c@{\hspace{2pt}}ccccc@{}}
    \rotatebox{90}{\scriptsize\ \ \ KITTI} &
    \includegraphics[width=0.188\linewidth]{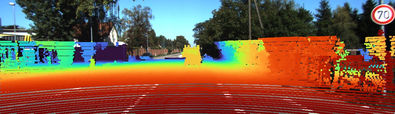} &
    \includegraphics[width=0.188\linewidth]{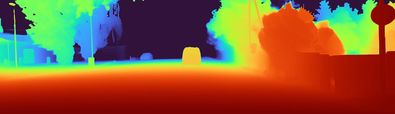} &
    \includegraphics[width=0.188\linewidth]{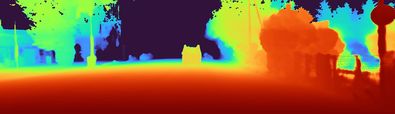} &
    \includegraphics[width=0.188\linewidth]{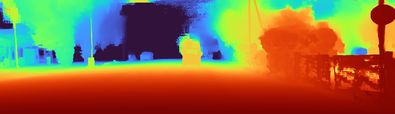} &
    \includegraphics[width=0.188\linewidth]{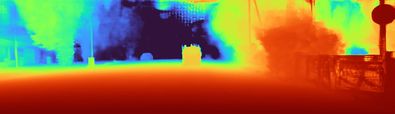} \\
    \rotatebox{90}{\scriptsize nuScenes} &
    \includegraphics[width=0.188\linewidth]{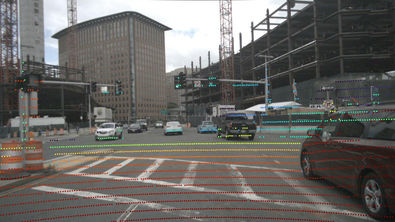} &
    \includegraphics[width=0.188\linewidth]{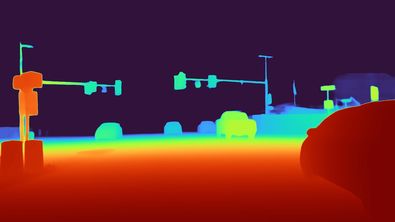} &
    \includegraphics[width=0.188\linewidth]{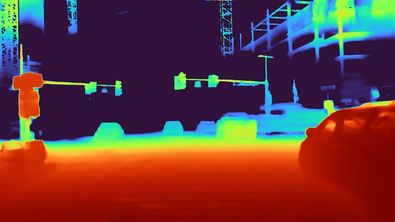} &
    \includegraphics[width=0.188\linewidth]{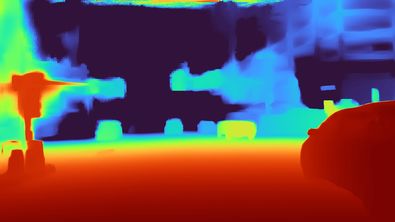} &
    \includegraphics[width=0.188\linewidth]{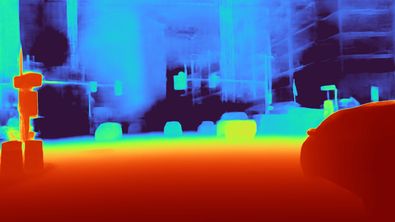} \\
    \rotatebox{90}{\scriptsize\ \ Waymo} &
    \includegraphics[width=0.188\linewidth]{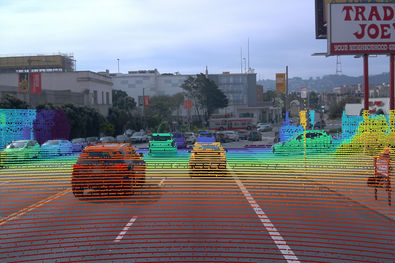} &
    \includegraphics[width=0.188\linewidth]{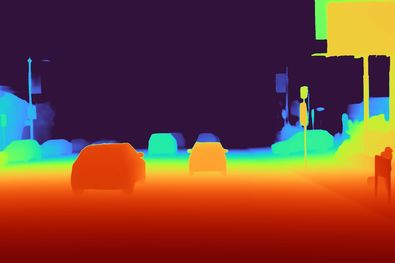} &
    \includegraphics[width=0.188\linewidth]{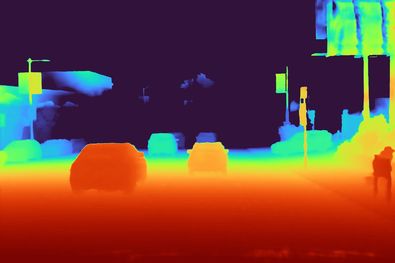} &
    \includegraphics[width=0.188\linewidth]{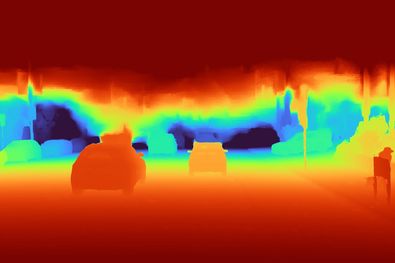} &
    \includegraphics[width=0.188\linewidth]{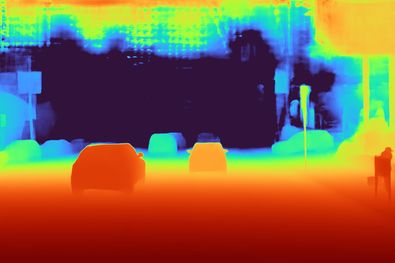} \\
    \rotatebox{90}{\scriptsize\ \ GOOSE} &
    \includegraphics[width=0.188\linewidth]{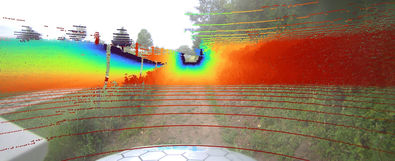} &
    \includegraphics[width=0.188\linewidth]{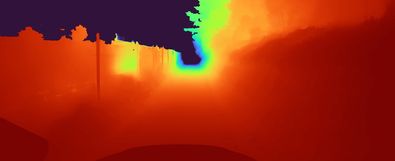} &
    \includegraphics[width=0.188\linewidth]{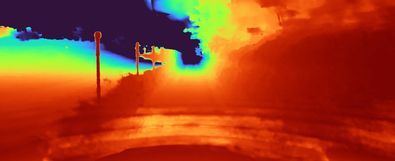} &
    \includegraphics[width=0.188\linewidth]{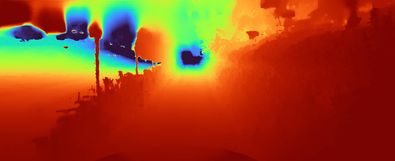} &
    \includegraphics[width=0.188\linewidth]{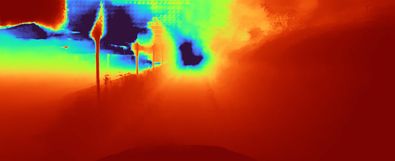} \\
    \rotatebox{90}{\scriptsize\ \ \ \ NYU} &
    \includegraphics[width=0.188\linewidth]{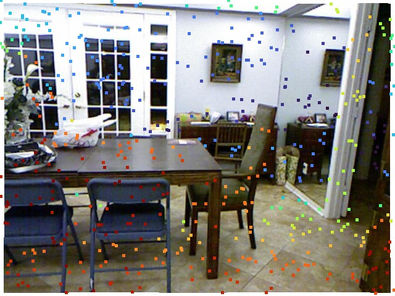} &
    \includegraphics[width=0.188\linewidth]{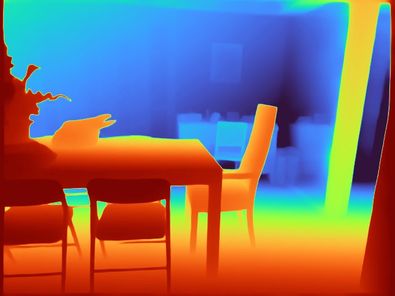} &
    \includegraphics[width=0.188\linewidth]{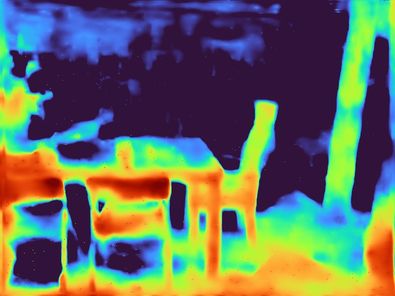} &
    \includegraphics[width=0.188\linewidth]{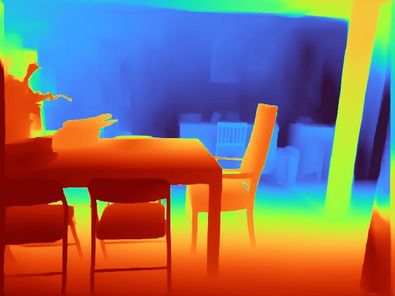} &
    \includegraphics[width=0.188\linewidth]{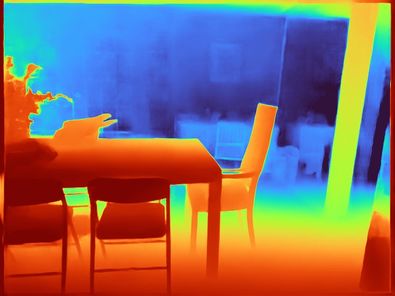} \\[3pt]
    & \scriptsize RGB + anchors & \scriptsize \textbf{ours} & \scriptsize DMD3C
    & \scriptsize OMNI-DC & \scriptsize Prior-DA \\
  \end{tabular}
  \caption{Qualitative bake-off across five datasets. Metric accuracy doesn't stand on its own, without appropriate runtime and coherent structure.}
  \label{fig:exp4_bakeoff_qual}
\end{figure*}

\section{Discussion and Limitations}

\begin{table}[t]
  \centering
  \setlength{\tabcolsep}{2.5pt}
  \renewcommand{\arraystretch}{0.95}
  \caption{Prior selection (KITTI \emph{val.}, $n{=}100$, A100, fp16). All rows: FounRef refined.
  Scored to 80\,m so each prior's bias beyond the 50\,m trust cap is visible; anchors stop at 50\,m,
  so the $>$50\,m column measures unsupported extrapolation. Structure is measured in log-depth on
  LiDAR-derived masks: \emph{incoh} = high-frequency spectral fraction on flat surfaces,
  \emph{detail} = the same on RGB-textured regions. Read them together: low+low = washed,
  high detail + low incoh = clean and sharp.}
  \label{tab:exp1_priors}
  \resizebox{\columnwidth}{!}{%
  \begin{tabular}{lccccccr}
    \toprule
    & \multicolumn{3}{c}{accuracy} & \multicolumn{2}{c}{structure} & \\
    \cmidrule(lr){2-4} \cmidrule(lr){5-6}
    prior & RMSE ($\downarrow$) & AbsRel ($\downarrow$) & RMSE$_{>50\text{m}}$ ($\downarrow$)
          & incoh ($\downarrow$) & detail ($\uparrow$) & ms ($\downarrow$) \\
    \midrule
    \textbf{MoGe-2}      & 2.83          & 0.029          & 14.57          & \textbf{0.007} & 0.025 & \textbf{49} \\
    Metric3D-V2          & \textbf{2.76} & \textbf{0.028} & 14.41          & 0.008          & 0.004 & 77 \\
    DA-V3                & 2.98          & 0.034          & 15.00          & 0.020          & 0.020 & 107 \\
    DA-V2                & 3.05          & 0.037          & 15.44          & 0.050          & \textbf{0.112} & 55 \\
    UniDepth-V2          & 3.59          & 0.047          & 17.28          & 0.020          & 0.007 & 49 \\
    DepthPro             & 2.83          & 0.032          & \textbf{14.38} & 0.213          & 0.035 & 205 \\
    \bottomrule
  \end{tabular}
  }
\end{table}
\begin{table}[t]
  \centering
  \scriptsize
  \setlength{\tabcolsep}{1.8pt}
  \renewcommand{\arraystretch}{0.95}
  \caption{Anchor-level filter comparison on KITTI validation ($n{=}100$).
  Errors are relative errors at dropped/kept anchors against dense GT or the
  independent reference $D_1$. Higher drop/keep ratios indicate better
  separation. Runtime is the filter's own cost on one A100 (fp16). Temporal is over 10 sweeps.The oracle measures the difference between the raw and filtered point clouds, as processed by the Authors of KITTI.}
  \label{tab:exp2_filter}
  \resizebox{\columnwidth}{!}{%
  \begin{tabular}{llcccccc}
    \toprule
    set & filter & drop & \multicolumn{2}{c}{vs.\ GT} & \multicolumn{2}{c}{vs.\ $D_1$} & ms \\
    \cmidrule(lr){4-5} \cmidrule(lr){6-7}
     & & (\%) & err.\ (D/K) & $\times$ & err.\ (D/K) & $\times$ & \\
    \midrule
    single & \textbf{Ours} & \textbf{2.3} & \textbf{1.08 / 0.04} & \textbf{27} & \textbf{1.10 / 0.03} & \textbf{37} & \textbf{13} \\
           & RePLAy        & 4.5 & 0.49 / 0.04 & 12 & 0.43 / 0.04 & 11 & 452 \\
           & Oracle        & 2.5 & 0.95 / 0.03 & 32 & 0.64 / 0.04 & 16 & --- \\
    \midrule
    temporal & \textbf{Ours} & \textbf{2.2} & \textbf{0.80 / 0.04} & \textbf{20} & \textbf{0.88 / 0.03} & \textbf{29} & \textbf{13} \\
             & RePLAy        & 2.2 & 0.26 / 0.04 & 7 & 0.27 / 0.05 & 6 & 868 \\
             & Oracle        & 1.5 & 0.87 / 0.03 & 29 & 0.59 / 0.05 & 12 & --- \\
    \bottomrule
  \end{tabular}%
  }
\end{table}

Our experiments show that dense metric depth should not be evaluated by point-wise error alone. Across prior selection, solver selection, and depth completion, lower RMSE can coincide with noisier surfaces, stronger anchor imprinting, or substantially higher runtime. These are distinct properties: metric accuracy determines whether the prediction is correctly grounded, structural fidelity determines whether the geometry remains coherent between sparse measurements, and runtime determines whether the method is practical for large-scale processing or online use. FounRef addresses this three-way objective by combining a transferable monocular prior, direct metric anchors, and lightweight structure-preserving refinement.

Table~\ref{tab:exp1_priors} identifies MoGe-2 as the most suitable prior for this balance. Metric3D-V2 achieves slightly lower refined metric error but preserves considerably less fine detail, whereas Depth Pro and Depth Anything introduce more high-frequency surface noise. MoGe-2 therefore provides the strongest overall combination of metric compatibility, coherent geometry, retained detail, and runtime. The fact that every tested prior improves beyond global calibration also indicates that FounRef is not specific to MoGe-2 and can directly benefit from future monocular foundation models.

The filtering and solver experiments clarify the remaining sources of error. FounRef removes only a small fraction of anchors, but the rejected measurements are substantially less reliable than those retained. Unlike geometry-only filtering, the surface-consistency test can reject cross-modal and temporal disagreements such as moving-object silhouettes, calibration errors, and occlusion artifacts. The solver comparison reveals a related trade-off: Poisson and TGV fit the anchors more aggressively and obtain lower metric error, but produce local bumps and visible anchor imprinting. The bilateral solver instead propagates corrections along surfaces defined by the prior and limits propagation across depth discontinuities, accepting a modest metric penalty in exchange for cleaner geometry and lower runtime.

The cross-domain benchmark exposes the limitations of learned depth completion and provides a diagnostic cross-section of current approaches. DMD3C is highly competitive on KITTI, but this advantage does not transfer consistently to nuScenes, Waymo, GOOSE, and NYU Depth V2. More generally, no learned baseline remains uniformly accurate across all five datasets. Supervised methods are sensitive to their training domains, sensor configurations, and sparse-depth patterns, while recent zero-shot and foundation-guided methods improve transfer at the cost of substantial runtime, surface degradation, or inconsistent metric accuracy. Consequently, none of the evaluated methods is an unambiguous out-of-the-box choice when accuracy, geometric coherence, cross-domain robustness, and efficiency are considered jointly.

FounRef behaves differently because it preserves the geometry of a frozen monocular prior and estimates only the metric correction supported by the sparse anchors. This separation allows it to inherit the prior's cross-domain structural generalization without learning a fixed RGB-sparse-depth mapping. As a result, FounRef achieves the lowest normal dispersion across all evaluated datasets while remaining metrically competitive and considerably faster than the learned alternatives. The more aggressive $\sigma_s{=}4$ configuration further improves metric accuracy without increasing runtime, but moves toward the same accuracy-structure trade-off observed in the solver study.

FounRef nevertheless remains limited by the quality of its prior. It can correct the metric placement of surfaces already represented by the monocular model, but cannot reliably recover geometry that the prior omits or interprets incorrectly. GOOSE illustrates this limitation: LiDAR may return through vegetation or fences, while the image prior represents the visible foreground as a coherent surface. FounRef tends to preserve that visible surface rather than create a sparse see-through reconstruction, which can increase point-wise error. Structure preservation is therefore an inductive bias rather than an unconditional advantage. Because FounRef is modular and directly inherits the geometry of its prior, stronger future priors could further improve both accuracy and structure, making the same lightweight framework an even stronger competitor to state-of-the-art depth-completion methods.

Overall, FounRef lies on an empirical Pareto frontier between metric accuracy, surface coherence, cross-domain robustness, and runtime. More aggressive local fitting can further reduce error, but increasingly exposes the sparse-anchor pattern and degrades surface quality, while still retaining substantially lower normal dispersion than the competing methods. A natural next step is a zero-shot refinement model that moves this frontier without sacrificing the structural generalization of the frozen prior or the efficiency of the current solver. Methods such as OMNI-DC and Prior-DA already move toward broader transfer and foundation-prior integration, but the benchmark shows that substantial gaps remain in surface coherence and runtime.
\section{Conclusion}

We introduced FounRef, a training-free method that converts a frozen monocular prior and sparse metric anchors into dense metric depth. Global calibration, surface-based anchor validation, and bilateral refinement provide metric grounding while preserving the prior's geometry. Across five datasets, FounRef remains metrically competitive with learned methods while producing more coherent surfaces at substantially lower runtime. Its modular design can directly benefit from stronger future priors and metric sensors without retraining.

\paragraph{Use of Generative AI.}
Generative AI was used for language editing. The authors verified all technical content, analyses, citations, and claims, and take full responsibility for the manuscript.

{
\bibliographystyle{plainnat}
\bibliography{references}

\begin{thebibliography}{26}
\providecommand{\natexlab}[1]{#1}
\providecommand{\url}[1]{\texttt{#1}}
\expandafter\ifx\csname urlstyle\endcsname\relax
  \providecommand{\doi}[1]{doi: #1}\else
  \providecommand{\doi}{doi: \begingroup \urlstyle{rm}\Url}\fi

\bibitem[Barron and Poole(2016)]{fastbilateralsolver}
Jonathan~T. Barron and Ben Poole.
\newblock {The Fast Bilateral Solver}.
\newblock In \emph{Proceedings of the European Conference on Computer Vision},
  pages 617--632, 2016.

\bibitem[Bochkovskii et~al.(2025)Bochkovskii, Delaunoy, Germain, Santos, Zhou,
  Richter, and Koltun]{depthpro}
Aleksei Bochkovskii, Ama{\"e}l Delaunoy, Hugo Germain, Marcel Santos, Yichao
  Zhou, Stephan~R. Richter, and Vladlen Koltun.
\newblock {Depth Pro: Sharp Monocular Metric Depth in Less Than a Second}.
\newblock In \emph{Proceedings of the International Conference on Learning
  Representations}, 2025.

\bibitem[Bredies et~al.(2010)Bredies, Kunisch, and Pock]{tgv}
Kristian Bredies, Karl Kunisch, and Thomas Pock.
\newblock {Total Generalized Variation}.
\newblock \emph{SIAM Journal on Imaging Sciences}, 3\penalty0 (3):\penalty0
  492--526, 2010.

\bibitem[Caesar et~al.(2020)Caesar, Bankiti, Lang, Vora, Liong, Xu, Krishnan,
  Pan, Baldan, and Beijbom]{nuscenes}
Holger Caesar, Varun Bankiti, Alex~H. Lang, Sourabh Vora, Venice~Erin Liong,
  Qiang Xu, Anush Krishnan, Yu~Pan, Giancarlo Baldan, and Oscar Beijbom.
\newblock {nuScenes: A Multimodal Dataset for Autonomous Driving}.
\newblock In \emph{Proceedings of the IEEE/CVF Conference on Computer Vision
  and Pattern Recognition}, pages 11621--11631, 2020.

\bibitem[Geiger et~al.(2013)Geiger, Lenz, Stiller, and Urtasun]{kitti}
Andreas Geiger, Philip Lenz, Christoph Stiller, and Raquel Urtasun.
\newblock {Vision Meets Robotics: The {KITTI} Dataset}.
\newblock \emph{The International Journal of Robotics Research}, 32\penalty0
  (11):\penalty0 1231--1237, 2013.

\bibitem[Guizilini et~al.(2023)Guizilini, Vasiljevic, Chen, Ambrus, and
  Gaidon]{zerodepth}
Vitor Guizilini, Igor Vasiljevic, Dian Chen, Rares Ambrus, and Adrien Gaidon.
\newblock {Towards Zero-Shot Scale-Aware Monocular Depth Estimation}.
\newblock In \emph{Proceedings of the IEEE/CVF International Conference on
  Computer Vision}, pages 9233--9243, 2023.

\bibitem[Hu et~al.(2024)Hu, Yin, Zhang, Cai, Long, Chen, Wang, Yu, Shen, and
  Shen]{metric3dv2}
Mu~Hu, Wei Yin, Chi Zhang, Zhipeng Cai, Xiaoxiao Long, Hao Chen, Kaixuan Wang,
  Gang Yu, Chunhua Shen, and Shaojie Shen.
\newblock {Metric3D v2: A Versatile Monocular Geometric Foundation Model for
  Zero-Shot Metric Depth and Surface Normal Estimation}.
\newblock \emph{arXiv preprint arXiv:2404.15506}, 2024.

\bibitem[Huber(1964)]{huber1964robust}
Peter~J. Huber.
\newblock {Robust Estimation of a Location Parameter}.
\newblock \emph{The Annals of Mathematical Statistics}, 35\penalty0
  (1):\penalty0 73--101, 1964.

\bibitem[Kazhdan and Hoppe(2013)]{screenedpoisson}
Michael Kazhdan and Hugues Hoppe.
\newblock {Screened Poisson Surface Reconstruction}.
\newblock \emph{ACM Transactions on Graphics}, 32\penalty0 (3):\penalty0
  29:1--29:13, 2013.

\bibitem[Liang et~al.(2025)Liang, Hu, Shao, and Fu]{dmd3c}
Yingping Liang, Yutao Hu, Wenqi Shao, and Ying Fu.
\newblock {Distilling Monocular Foundation Model for Fine-grained Depth
  Completion}.
\newblock In \emph{Proceedings of the IEEE/CVF Conference on Computer Vision
  and Pattern Recognition}, 2025.

\bibitem[Lin et~al.(2025{\natexlab{a}})Lin, Chen, Liew, Chen, Li, Shi, Feng,
  and Kang]{lin2025depthanything3}
Haotong Lin, Sili Chen, Jun~Hao Liew, Donny~Y. Chen, Zhenyu Li, Guang Shi,
  Jiashi Feng, and Bingyi Kang.
\newblock {Depth Anything 3: Recovering the Visual Space from Any Views}.
\newblock \emph{arXiv preprint arXiv:2511.10647}, 2025{\natexlab{a}}.

\bibitem[Lin et~al.(2025{\natexlab{b}})Lin, Peng, Chen, Peng, Sun, Liu, Bao,
  Feng, Zhou, and Kang]{lin2025promptda}
Haotong Lin, Sida Peng, Jingxiao Chen, Songyou Peng, Jiaming Sun, Minghuan Liu,
  Hujun Bao, Jiashi Feng, Xiaowei Zhou, and Bingyi Kang.
\newblock {Prompting Depth Anything for 4K Resolution Accurate Metric Depth
  Estimation}.
\newblock In \emph{Proceedings of the IEEE/CVF Conference on Computer Vision
  and Pattern Recognition}, 2025{\natexlab{b}}.

\bibitem[Ma and Karaman(2018)]{ma2018sparse2dense}
Fangchang Ma and Sertac Karaman.
\newblock {Sparse-to-Dense: Depth Prediction from Sparse Depth Samples and a
  Single Image}.
\newblock In \emph{Proceedings of the IEEE International Conference on Robotics
  and Automation}, pages 4796--4803, 2018.

\bibitem[Mortimer et~al.(2024)Mortimer, Hagmanns, Granero, Luettel, Petereit,
  and Wuensche]{goose}
Peter Mortimer, Raphael Hagmanns, Miguel Granero, Thorsten Luettel, Janko
  Petereit, and Hans-Joachim Wuensche.
\newblock {The {GOOSE} Dataset for Perception in Unstructured Environments}.
\newblock In \emph{Proceedings of the IEEE International Conference on Robotics
  and Automation}, 2024.

\bibitem[Park et~al.(2020)Park, Joo, Hu, Liu, and Kweon]{nlspn}
Jinsun Park, Kyungdon Joo, Zhe Hu, Chi-Kuei Liu, and In~So Kweon.
\newblock {Non-Local Spatial Propagation Network for Depth Completion}.
\newblock In \emph{Proceedings of the European Conference on Computer Vision},
  2020.

\bibitem[Piccinelli et~al.(2024)Piccinelli, Yang, Sakaridis, Segu, Li,
  Van~Gool, and Yu]{unidepth}
Luigi Piccinelli, Yung-Hsu Yang, Christos Sakaridis, Mattia Segu, Siyuan Li,
  Luc Van~Gool, and Fisher Yu.
\newblock {UniDepth: Universal Monocular Metric Depth Estimation}.
\newblock In \emph{Proceedings of the IEEE/CVF Conference on Computer Vision
  and Pattern Recognition}, 2024.

\bibitem[Silberman et~al.(2012)Silberman, Hoiem, Kohli, and Fergus]{nyuv2}
Nathan Silberman, Derek Hoiem, Pushmeet Kohli, and Rob Fergus.
\newblock {Indoor Segmentation and Support Inference from {RGBD} Images}.
\newblock In \emph{Proceedings of the European Conference on Computer Vision},
  pages 746--760, 2012.

\bibitem[Sun et~al.(2020)Sun, Kretzschmar, Dotiwalla, Chouard, Patnaik, Tsui,
  Guo, Zhou, Chai, Caine, Vasudevan, Han, Ngiam, Zhao, Timofeev, Ettinger,
  Krivokon, Gao, Joshi, Zhao, Cheng, Zhang, Shlens, Chen, and Anguelov]{waymo}
Pei Sun, Henrik Kretzschmar, Xerxes Dotiwalla, Aurelien Chouard, Vijaysai
  Patnaik, Paul Tsui, James Guo, Yin Zhou, Yuning Chai, Benjamin Caine, Vijay
  Vasudevan, Wei Han, Jiquan Ngiam, Hang Zhao, Aleksei Timofeev, Scott
  Ettinger, Maxim Krivokon, Amy Gao, Aditya Joshi, Sheng Zhao, Shuyang Cheng,
  Yu~Zhang, Jonathon Shlens, Zhifeng Chen, and Dragomir Anguelov.
\newblock {Scalability in Perception for Autonomous Driving: Waymo Open
  Dataset}.
\newblock In \emph{Proceedings of the IEEE/CVF Conference on Computer Vision
  and Pattern Recognition}, pages 2446--2454, 2020.

\bibitem[Tang et~al.(2024)Tang, Tian, An, Li, and Tan]{tang2024bpnet}
Jie Tang, Fei-Peng Tian, Boshi An, Jian Li, and Ping Tan.
\newblock {Bilateral Propagation Network for Depth Completion}.
\newblock In \emph{Proceedings of the IEEE/CVF Conference on Computer Vision
  and Pattern Recognition}, pages 9763--9772, 2024.

\bibitem[Uhrig et~al.(2017)Uhrig, Schneider, Schneider, Franke, Brox, and
  Geiger]{uhrig2017sparsity}
Jonas Uhrig, Nick Schneider, Lukas Schneider, Uwe Franke, Thomas Brox, and
  Andreas Geiger.
\newblock {Sparsity Invariant {CNNs}}.
\newblock In \emph{Proceedings of the International Conference on 3D Vision},
  pages 11--20, 2017.

\bibitem[Wang et~al.(2025{\natexlab{a}})Wang, Xu, Dong, Deng, Xiang, Lv, Sun,
  Tong, and Yang]{moge2}
Ruicheng Wang, Sicheng Xu, Yue Dong, Yu~Deng, Jianfeng Xiang, Zelong Lv,
  Guangzhong Sun, Xin Tong, and Jiaolong Yang.
\newblock {MoGe-2: Accurate Monocular Geometry with Metric Scale and Sharp
  Details}.
\newblock \emph{arXiv preprint arXiv:2507.02546}, 2025{\natexlab{a}}.

\bibitem[Wang et~al.(2025{\natexlab{b}})Wang, Chen, Yang, Wang, Zhang, Zhao,
  and Zhao]{wang2025priorda}
Zehan Wang, Siyu Chen, Lihe Yang, Jialei Wang, Ziang Zhang, Hengshuang Zhao,
  and Zhou Zhao.
\newblock {Depth Anything with Any Prior}.
\newblock \emph{arXiv preprint arXiv:2505.10565}, 2025{\natexlab{b}}.

\bibitem[Yang et~al.(2024)Yang, Kang, Huang, Zhao, Xu, Feng, and
  Zhao]{depthanythingv2}
Lihe Yang, Bingyi Kang, Zilong Huang, Zhen Zhao, Xiaogang Xu, Jiashi Feng, and
  Hengshuang Zhao.
\newblock {Depth Anything V2}.
\newblock In \emph{Advances in Neural Information Processing Systems}, 2024.

\bibitem[Zhang et~al.(2023)Zhang, Guo, Poggi, Zhu, Huang, and
  Mattoccia]{completionformer}
Youmin Zhang, Xianda Guo, Matteo Poggi, Zheng Zhu, Guan Huang, and Stefano
  Mattoccia.
\newblock {CompletionFormer: Depth Completion with Convolutions and Vision
  Transformers}.
\newblock In \emph{Proceedings of the IEEE/CVF Conference on Computer Vision
  and Pattern Recognition}, pages 18527--18536, 2023.

\bibitem[Zhu et~al.(2024)Zhu, Ganesan, Kumar, and Liu]{replay2024}
Shengjie Zhu, Girish~Chandar Ganesan, Abhinav Kumar, and Xiaoming Liu.
\newblock {{RePLAy}: Remove Projective {LiDAR} Depthmap Artifacts via
  Exploiting Epipolar Geometry}.
\newblock \emph{arXiv preprint arXiv:2407.19154}, 2024.

\bibitem[Zuo et~al.(2025)Zuo, Yang, Ma, and Deng]{omnidc}
Yiming Zuo, Willow Yang, Zeyu Ma, and Jia Deng.
\newblock {{OMNI-DC}: Highly Robust Depth Completion with Multiresolution Depth
  Integration}.
\newblock In \emph{Proceedings of the IEEE/CVF International Conference on
  Computer Vision}, 2025.

\end{thebibliography}
}

\appendix
\section{Appendix}
\label{sec:appendix}

This appendix ablates the pipeline stage by stage, reports how the method behaves as the metric
evidence is thinned and corrupted, examines the two solver settings that admit a continuous choice, and
compares every method qualitatively across the five datasets.

Every number reported here uses the default configuration; nothing was retuned for the appendix.
These experiments use their own frame subset and reference for screening the evaluation target, so
absolute values are not directly comparable with the main tables --- each table is internally consistent,
and the comparisons drawn are within a table.

\paragraph{Pipeline break-down. }
\label{sec:apx-stages-sec}

\begin{figure}[!ht]
\centering
\begin{tabular}{@{}c@{\hspace{2pt}}c@{}}
\includegraphics[width=0.48\columnwidth]{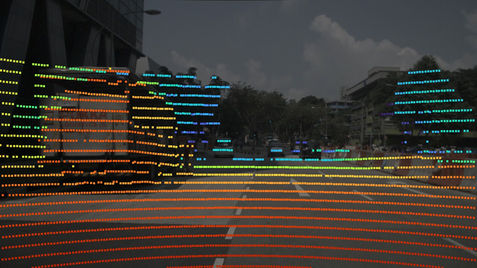} & \includegraphics[width=0.48\columnwidth]{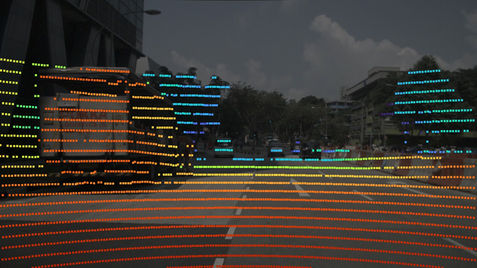} \\
\small anchors in & \small kept, step 5 \\[3pt]
\includegraphics[width=0.48\columnwidth]{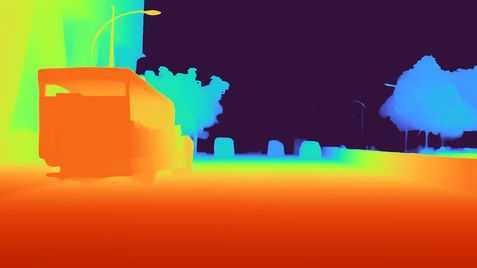} & \includegraphics[width=0.48\columnwidth]{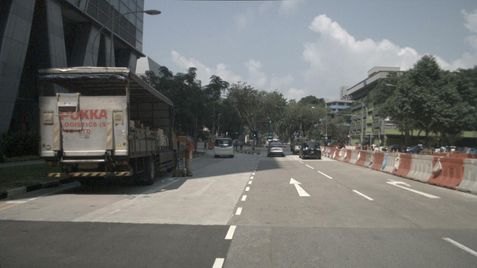} \\
\small $D_{\mathrm{prior}}$ (monocular) & \small input RGB \\[3pt]
\includegraphics[width=0.48\columnwidth]{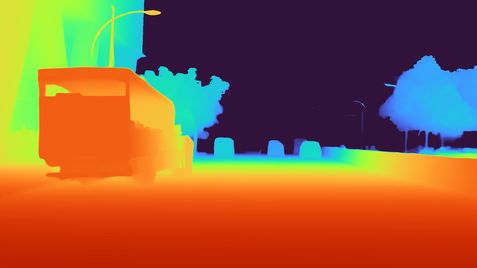} & \includegraphics[width=0.48\columnwidth]{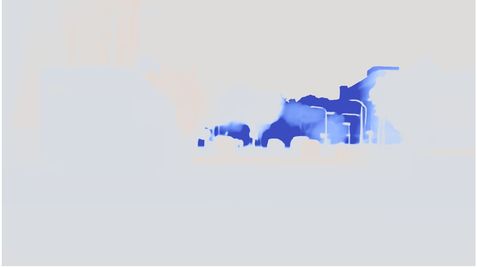} \\
\small $D_{\mathrm{cal}}$ (calibrated) & \small $\Delta$ calibrate \\[3pt]
\includegraphics[width=0.48\columnwidth]{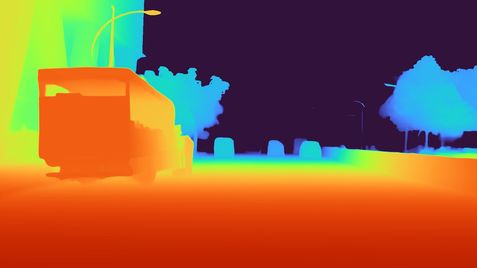} & \includegraphics[width=0.48\columnwidth]{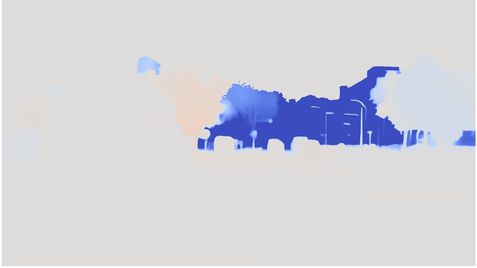} \\
\small $D_1$ (first solve) & \small $\Delta$ solve $D_1$ \\[3pt]
\includegraphics[width=0.48\columnwidth]{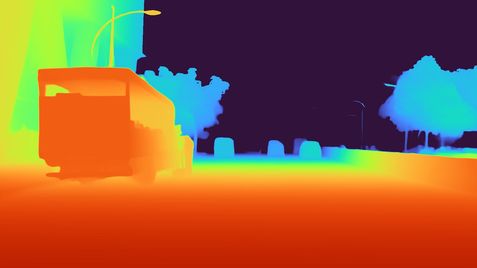} & \includegraphics[width=0.48\columnwidth]{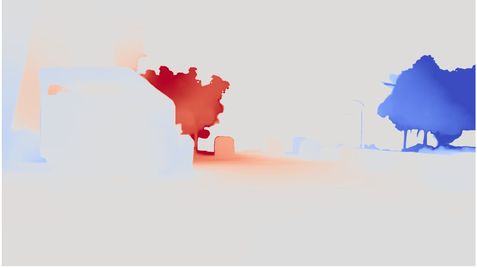} \\
\small $D_2$ (second solve) & \small $\Delta$ solve $D_2$ \\
\end{tabular}
\caption{The pipeline on one nuScenes frame, by row: left the state after a step,
right what that step changed (blue nearer, red farther, grey unchanged, each panel on its own scale).
Top: anchors in, and what survives the anchor filter ($4.7\%$ removed), dilated for legibility.}
\label{fig:apx-strip}
\end{figure}
\begin{table}[!ht]
\centering
\setlength{\tabcolsep}{4pt}
\footnotesize
\begin{tabular}{lrrr}
\toprule
stage & \textbf{kitti} & \textbf{waymo} & \textbf{goose} \\
\midrule
\multicolumn{4}{l}{\emph{RMSE (m)}} \\
\quad calibrated prior & 1.407 & 2.522 & 3.268 \\
\quad $+\,D_1$ (filter reference) & 0.698 & 0.959 & 1.101 \\
\quad $+\,D_2$ (full method) & 0.567 & 0.826 & 1.131 \\
\multicolumn{4}{l}{\emph{RMSE 30--50\,m}} \\
\quad calibrated prior & 4.220 & 4.423 & 8.713 \\
\quad $+\,D_1$ (filter reference) & 2.021 & 1.836 & 2.916 \\
\quad $+\,D_2$ (full method) & 1.636 & 1.581 & 3.032 \\
\multicolumn{4}{l}{\emph{normal disp. (deg)}} \\
\quad calibrated prior & 10.64 & 3.15 & 8.94 \\
\quad $+\,D_1$ (filter reference) & 11.65 & 5.40 & 9.37 \\
\quad $+\,D_2$ (full method) & 11.78 & 4.83 & 9.54 \\
\bottomrule
\end{tabular}
\caption{Pipeline ablation on the accumulated, \emph{unfiltered} cloud. Anchors and
held-out target are screened by the method's own MoGe reference; no competitor appears here, so these are
inference numbers rather than the cross-method protocol of the main tables.}
\label{tab:apx-stages}
\end{table}

Table~\ref{tab:apx-stages} adds one stage of the pipeline at a time on KITTI, Waymo and GOOSE, using
the cloud \emph{before} any cleaning so the noise the anchor filter exists for is present. Global
calibration fixes the metric scale of the monocular prior; the first solve $D_1$ produces the surface the
filter tests each anchor against; the second solve $D_2$ runs on the anchors that survive it.

Each stage pays for itself. The first solve $D_1$ removes half to two thirds of the calibrated prior's error, and
$D_2$ improves on $D_1$ by $19\%$ on KITTI and $14\%$ on Waymo --- that margin is what filtering the
cloud buys, since $D_1$ and $D_2$ are the same solver differing in the anchors they receive. GOOSE is the
exception at $2.7\%$ the other way: off-road returns through vegetation disagree with the image-visible
surface for reasons the threshold cannot separate from noise.

Figure~\ref{fig:apx-strip} shows the whole progression on one frame: each stage in turn, the anchor
cloud the filter accepts and rejects, and the change every step applies. It is the visual counterpart of the dispersion column --- what the stages move is the metric
field, not the geometry.

\paragraph{Behaviour as anchors are thinned and mis-projected. }
\label{sec:apx-density}

\begin{figure}[!ht]
\centering
\includegraphics[width=\columnwidth]{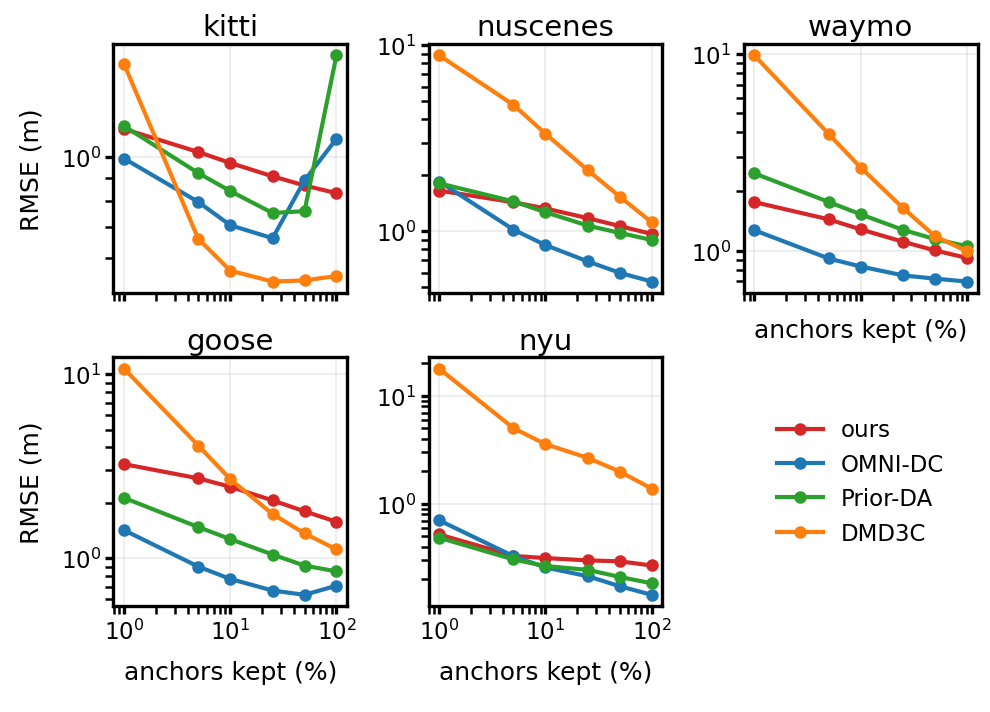}
\caption{Anchor-density sweep.}
\label{fig:apx-density}
\end{figure}

How many anchors a frame carries is a property of the rig, not of the method: a $32$-beam scanner
returns roughly half what a $64$-beam one does, returns thin out with range and on dark or specular
surfaces, and accumulation is available only when ego-motion is. Figure~\ref{fig:apx-density} therefore sweeps the
retained fraction from $100\%$ to $1\%$, with every method receiving the identical thinned cloud on each
frame and scored against the unchanged target.

\begin{table}[!ht]
\centering
\setlength{\tabcolsep}{2.5pt}
\footnotesize
\begin{tabular}{lrrrrrr}
\toprule
method & 100\% & 50\% & 25\% & 10\% & 5\% & 1\% \\
\midrule
\multicolumn{7}{l}{\emph{kitti}} \\
\quad ours & 0.832 & 0.865 & 0.906 & 0.969 & 1.024 & 1.150 \\
\quad OMNI-DC & 1.095 & 0.887 & 0.663 & 0.708 & 0.796 & 0.991 \\
\quad Prior-Depth-Anything & 1.667 & 0.760 & 0.753 & 0.841 & 0.921 & 1.167 \\
\quad DMD3C & 0.548 & 0.536 & 0.533 & 0.563 & 0.661 & 1.596 \\
\multicolumn{7}{l}{\emph{nuscenes}} \\
\quad ours & 0.968 & 1.067 & 1.177 & 1.327 & 1.433 & 1.655 \\
\quad OMNI-DC & 0.537 & 0.598 & 0.689 & 0.843 & 1.021 & 1.852 \\
\quad Prior-Depth-Anything & 0.898 & 0.981 & 1.073 & 1.266 & 1.448 & 1.811 \\
\quad DMD3C & 1.118 & 1.537 & 2.131 & 3.347 & 4.784 & 8.832 \\
\multicolumn{7}{l}{\emph{waymo}} \\
\quad ours & 0.917 & 1.001 & 1.109 & 1.282 & 1.442 & 1.766 \\
\quad OMNI-DC & 0.695 & 0.719 & 0.747 & 0.829 & 0.911 & 1.269 \\
\quad Prior-Depth-Anything & 1.052 & 1.144 & 1.275 & 1.530 & 1.769 & 2.473 \\
\quad DMD3C & 0.992 & 1.184 & 1.650 & 2.649 & 3.928 & 9.892 \\
\multicolumn{7}{l}{\emph{goose}} \\
\quad ours & 1.578 & 1.800 & 2.062 & 2.445 & 2.718 & 3.247 \\
\quad OMNI-DC & 0.710 & 0.633 & 0.668 & 0.772 & 0.901 & 1.421 \\
\quad Prior-Depth-Anything & 0.849 & 0.912 & 1.048 & 1.271 & 1.477 & 2.129 \\
\quad DMD3C & 1.117 & 1.365 & 1.739 & 2.690 & 4.097 & 10.720 \\
\multicolumn{7}{l}{\emph{nyu}} \\
\quad ours & 0.269 & 0.294 & 0.300 & 0.315 & 0.330 & 0.521 \\
\quad OMNI-DC & 0.144 & 0.173 & 0.214 & 0.258 & 0.328 & 0.704 \\
\quad Prior-Depth-Anything & 0.184 & 0.211 & 0.245 & 0.266 & 0.306 & 0.486 \\
\quad DMD3C & 1.385 & 1.989 & 2.666 & 3.566 & 5.016 & 17.832 \\
\bottomrule
\end{tabular}
\caption{Anchor-density sweep: RMSE (m) as a fraction of the returns is withheld. Every method
receives the identical thinned cloud on each frame, drawn once per frame and shared, and is scored against
the same target. $100\%$ is the temporally accumulated cloud: $135\,351$ anchors per frame on KITTI
($31.6\%$ of all pixels), $35\,815$ on nuScenes ($2.5\%$), $109\,119$ on Waymo ($4.4\%$) and $102\,247$ on
GOOSE ($4.2\%$).}
\label{tab:apx-density-tab}
\end{table}

Table~\ref{tab:apx-density-tab} gives the numbers behind the curves. The methods separate by how
they degrade, not by where they start. From full density to $1\%$ our
error grows by $1.4$--$2.1\times$ on every dataset, the most uniform response in the table,
because the prior supplies the geometry and the anchors only fix its metric scale --- removing anchors
weakens the scale estimate without touching the surface. DMD3C, which regresses depth from the points
themselves, grows by $2.9$--$12.9\times$ and is unusable at the sparse end.

Two baselines are non-monotone: OMNI-DC improves from $1.095$ to $0.663$ on KITTI and from
$0.710$ to $0.633$ on GOOSE as the cloud is thinned, and Prior-Depth-Anything from $1.667$ to $0.753$ on
KITTI. This is not an artefact of the sweep --- the thinned subset is drawn once per frame and given to
every method alike, and our own curve is monotone on all five datasets. It follows from how those two
clouds are accumulated. KITTI is registered by scan-to-scan ICP, which drifts, and GOOSE is accumulated from ego-motion alone, so anything that moves smears along its trajectory; nuScenes and Waymo use dataset ego-poses with
per-object motion compensation, and neither shows the effect. Accumulation on KITTI and GOOSE
therefore injects anchors that are individually wrong. A completion network propagates
each measurement almost as a hard constraint, so a wrong anchor writes a wrong depth into its
neighbourhood, and thinning removes a share of them and lets the image evidence take over. Our solve is
affected far less because a bilateral cell pools several anchors --- $3.9$ per vertex at the default
$\sigma_s$ --- so no single measurement moves the field.

\begin{table*}[!tbp]
\centering
\setlength{\tabcolsep}{5pt}
\small
\begin{tabular}{lrrrrrrrrrr}
\toprule
$\sigma_s$ & \multicolumn{2}{c}{KITTI} & \multicolumn{2}{c}{nuScenes} & \multicolumn{2}{c}{Waymo} & \multicolumn{2}{c}{GOOSE} & \multicolumn{2}{c}{NYU} \\
\cmidrule(lr){2-3}\cmidrule(lr){4-5}\cmidrule(lr){6-7}\cmidrule(lr){8-9}\cmidrule(lr){10-11}
 & RMSE & disp. & RMSE & disp. & RMSE & disp. & RMSE & disp. & RMSE & disp. \\
\midrule
16 (default) & 0.832 & 12.01 & 0.968 & 6.94 & 0.917 & 4.85 & 2.254 & 9.35 & 0.269 & 7.53 \\
12 & 0.794 & 11.88 & 0.908 & 6.84 & 0.859 & 4.80 & 2.043 & 9.29 & 0.264 & 7.53 \\
8 & 0.736 & 11.71 & 0.826 & 6.72 & 0.791 & 4.77 & 1.798 & 9.18 & 0.258 & 7.53 \\
6 & 0.700 & 11.60 & 0.776 & 6.65 & 0.748 & 4.78 & 1.651 & 9.18 & 0.255 & 7.53 \\
4 & 0.656 & 11.49 & 0.713 & 6.61 & 0.704 & 4.85 & 1.513 & 9.36 & 0.253 & 7.53 \\
3 & 0.630 & 11.64 & 0.677 & 6.67 & 0.683 & 5.06 & 1.443 & 9.75 & 0.253 & 7.55 \\
2 & 0.607 & 12.37 & 0.655 & 7.06 & 0.674 & 6.00 & 1.365 & 11.18 & 0.258 & 7.60 \\
\bottomrule
\end{tabular}
\caption{Solver operating points: the bilateral cell $\sigma_s$ swept at constant $\lambda$, over
50 frames per dataset. RMSE in metres and normal dispersion in degrees side by side, so the
accuracy--structure trade of each setting reads across a row.}
\label{tab:apx-operating}
\end{table*}
\begin{table*}[!tbp]
\centering
\setlength{\tabcolsep}{5pt}
\small
\begin{tabular}{lrrrrrrrrrrr}
\toprule
variant & \multicolumn{2}{c}{KITTI} & \multicolumn{2}{c}{nuScenes} & \multicolumn{2}{c}{Waymo} & \multicolumn{2}{c}{GOOSE} & \multicolumn{2}{c}{NYU} & ms \\
\cmidrule(lr){2-3}\cmidrule(lr){4-5}\cmidrule(lr){6-7}\cmidrule(lr){8-9}\cmidrule(lr){10-11}
 & RMSE & disp. & RMSE & disp. & RMSE & disp. & RMSE & disp. & RMSE & disp. & \\
\midrule
default configuration & 0.832 & 12.01 & 0.968 & 6.94 & 0.917 & 4.85 & 2.254 & 9.35 & 0.269 & 7.53 & 35 \\
full-resolution solve & 0.793 & 11.52 & 1.021 & 6.58 & 0.931 & 4.62 & 2.622 & 9.35 & 0.282 & 7.50 & 40 \\
CG 2 iterations & 0.868 & 11.25 & 1.195 & 6.54 & 1.074 & 4.46 & 3.744 & 9.69 & 0.293 & 7.48 & 27 \\
CG 16 iterations & 0.831 & 12.16 & 0.961 & 7.27 & 0.916 & 5.02 & 2.327 & 9.44 & 0.270 & 7.52 & 42 \\
bistochastic 8 steps & 0.832 & 11.97 & 0.967 & 6.92 & 0.917 & 4.84 & 2.253 & 9.20 & 0.269 & 7.53 & 35 \\
\bottomrule
\end{tabular}
\caption{Solver components, each toggled alone from the default configuration, over 50
frames per dataset; RMSE in metres and normal dispersion in degrees side by side, `ms' the mean solve
time. \emph{Full-resolution solve} drops the half-resolution grid. \emph{CG iterations} is the conjugate-gradient
budget of the final solve. \emph{Bistochastic normalisation} is the grid balancing that makes the
bilateral operator well conditioned. A component is worth keeping only if switching it off costs more
than the runtime it saves.}
\label{tab:apx-parts}
\end{table*}

\begin{table}[!ht]
\centering
\setlength{\tabcolsep}{4pt}
\footnotesize
\begin{tabular}{lrrrrr}
\toprule
perturbation & \textbf{kitti} & \textbf{nuscenes} & \textbf{waymo} & \textbf{goose} & \textbf{nyu} \\
\midrule
\multicolumn{6}{l}{\emph{projection shift}} \\
\quad 0\,px & 0.832 & 0.968 & 0.917 & 2.254 & 0.269 \\
\quad 1\,px & 0.911 & 0.969 & 0.932 & 2.251 & 0.269 \\
\quad 2\,px & 1.042 & 1.003 & 0.972 & 2.254 & 0.270 \\
\quad 3\,px & 1.219 & 1.043 & 1.033 & 2.264 & 0.272 \\
\quad 5\,px & 1.624 & 1.151 & 1.178 & 2.293 & 0.274 \\
\multicolumn{6}{l}{\emph{multiplicative depth noise}} \\
\quad 0\% & 0.832 & 0.968 & 0.917 & 2.254 & 0.269 \\
\quad 1\% & 0.834 & 0.968 & 0.919 & 2.255 & 0.268 \\
\quad 2\% & 0.836 & 0.969 & 0.922 & 2.253 & 0.268 \\
\quad 5\% & 0.850 & 0.978 & 0.945 & 2.250 & 0.266 \\
\bottomrule
\end{tabular}
\caption{Anchor perturbation, ours. Only the anchors are corrupted; the evaluation target is the unperturbed held-out cloud (dense GT on KITTI/NYU), so the numbers measure damage to the prediction rather than a moved target.}
\label{tab:apx-perturb}
\end{table}

Table~\ref{tab:apx-perturb} corrupts the anchors instead of removing them. Only the input is
perturbed; the evaluation target is the unperturbed
held-out cloud, so the numbers measure damage to the prediction rather than a moved target. Depth noise
is comparatively benign --- a $5\%$ multiplicative error moves every dataset by about $3\%$ or less ---
whereas projection error is not: a $5$-pixel shift raises KITTI error by $95\%$. FounRef is thus robust
to moderate depth noise but remains sensitive to extrinsic or timing miscalibration, motivating accurate
sensor synchronisation.

\paragraph{Solver sensitivity. }
\label{sec:apx-solver}

The solver exposes two continuous settings, the bilateral cell $\sigma_s$ and the smoothness weight
$\lambda$, and a handful of discrete ones --- solve resolution, conjugate-gradient budget, bistochastic
steps --- which Table~\ref{tab:apx-parts} ablates separately. Table~\ref{tab:apx-operating} sweeps
$\sigma_s$ at fixed $\lambda$ around the default configuration and reports the accuracy--structure pair
each setting produces.

Neither axis is fragile. Normal dispersion moves by a few percent across the whole range
from $\sigma_s\!=\!16$ down to $3$, an order of magnitude in cell size, and degrades only at
$\sigma_s\!=\!2$ (Waymo $4.85\to6.00$, GOOSE $9.35\to11.18$), where the cell is too fine for neighbouring
anchors to be pooled at all. Accuracy, meanwhile, improves monotonically as the cell shrinks: the
default $\sigma_s\!=\!16$ is the least accurate setting in the sweep, giving up $21$--$33\%$ of its RMSE
to $\sigma_s\!=\!4$ on the four outdoor datasets and $6\%$ on NYU, and the frame-averaged dispersion does
not object.

\begin{figure*}[!tbp]
\centering
\begin{tabular}{@{}c@{\hspace{1.5pt}}c@{\hspace{1.5pt}}c@{\hspace{1.5pt}}c@{}}
\includegraphics[width=0.235\linewidth]{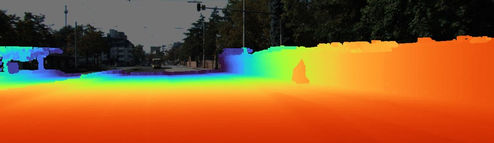} & \includegraphics[width=0.235\linewidth]{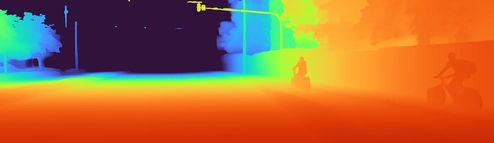} & \includegraphics[width=0.235\linewidth]{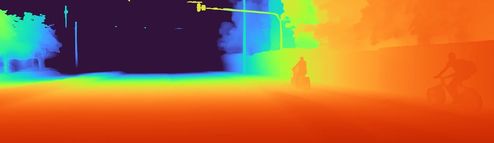} & \includegraphics[width=0.235\linewidth]{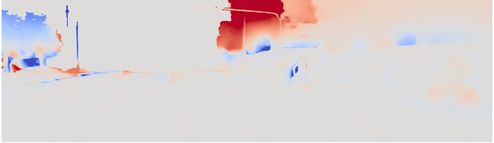} \\
\small KITTI: anchors in & \small $\sigma_s\!=\!16$ & \scriptsize $\sigma_s\!=\!4$ & \scriptsize difference \\
\includegraphics[width=0.235\linewidth]{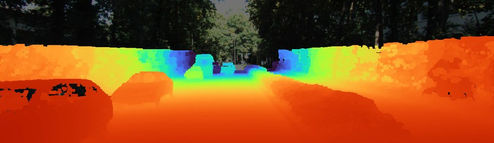} & \includegraphics[width=0.235\linewidth]{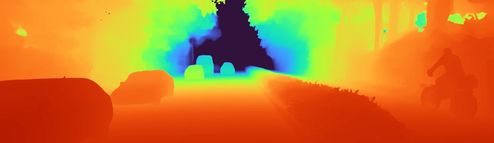} & \includegraphics[width=0.235\linewidth]{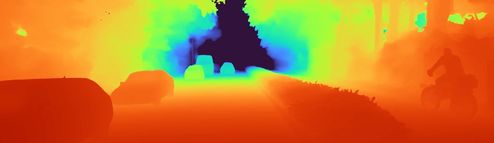} & \includegraphics[width=0.235\linewidth]{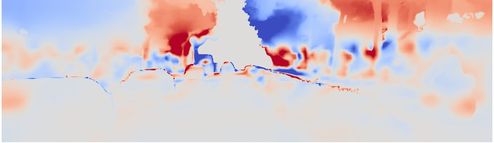} \\
\small KITTI: anchors in & \small $\sigma_s\!=\!16$ & \scriptsize $\sigma_s\!=\!4$ & \scriptsize difference \\
\includegraphics[width=0.235\linewidth]{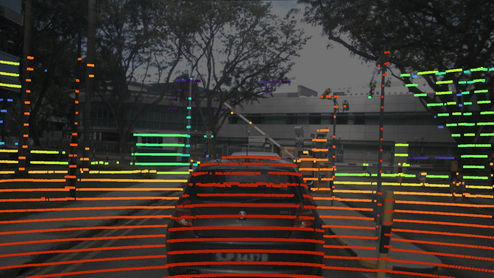} & \includegraphics[width=0.235\linewidth]{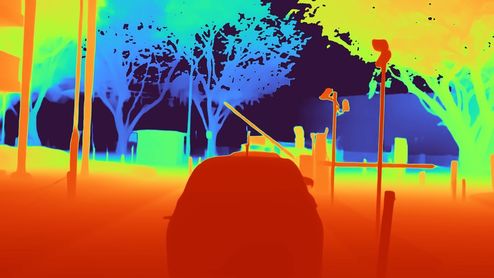} & \includegraphics[width=0.235\linewidth]{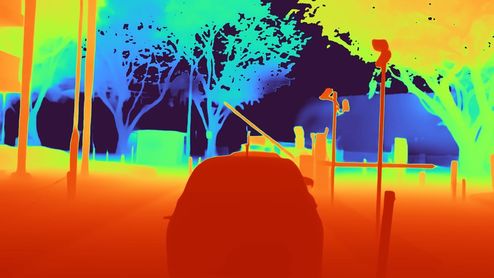} & \includegraphics[width=0.235\linewidth]{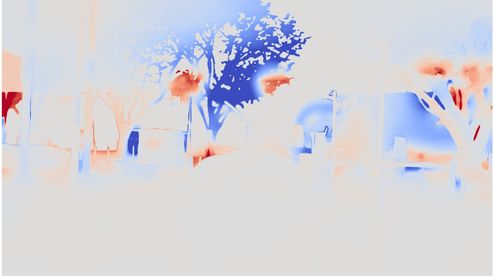} \\
\small nuScenes: anchors in & \small $\sigma_s\!=\!16$ & \scriptsize $\sigma_s\!=\!4$ & \scriptsize difference \\
\includegraphics[width=0.235\linewidth]{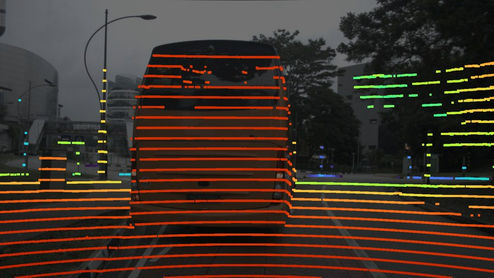} & \includegraphics[width=0.235\linewidth]{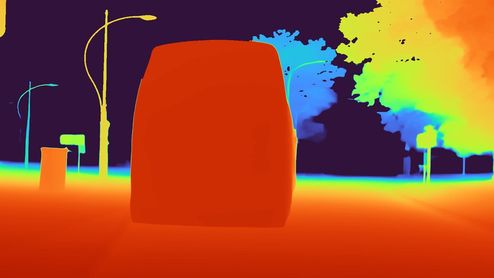} & \includegraphics[width=0.235\linewidth]{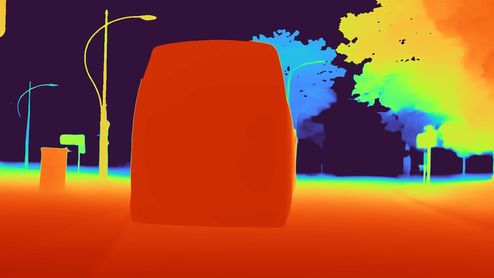} & \includegraphics[width=0.235\linewidth]{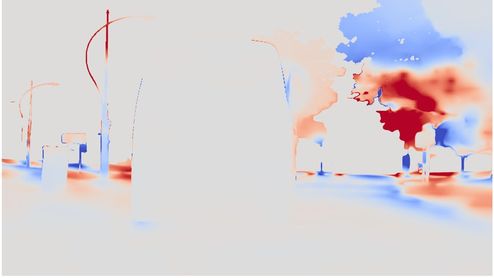} \\
\small nuScenes: anchors in & \small $\sigma_s\!=\!16$ & \scriptsize $\sigma_s\!=\!4$ & \scriptsize difference \\
\includegraphics[width=0.235\linewidth]{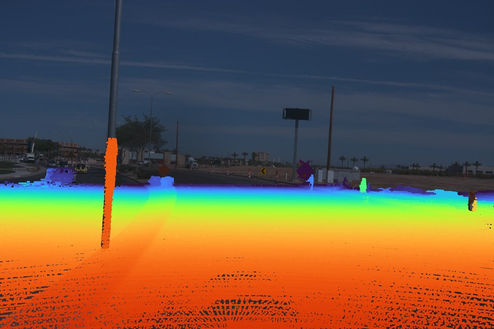} & \includegraphics[width=0.235\linewidth]{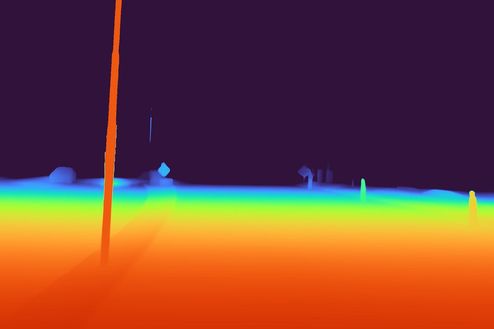} & \includegraphics[width=0.235\linewidth]{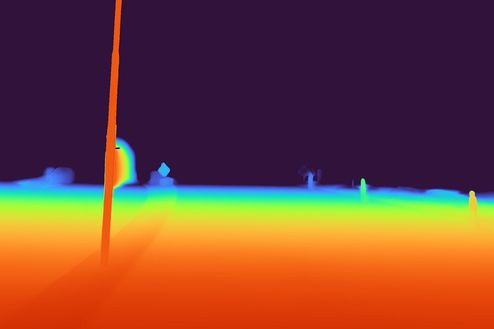} & \includegraphics[width=0.235\linewidth]{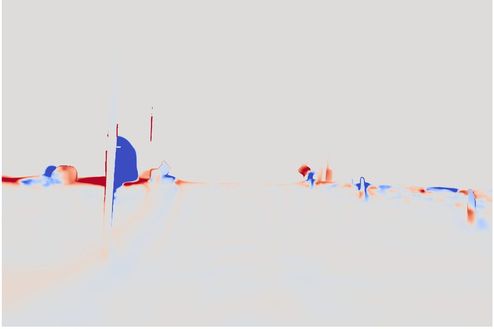} \\
\small Waymo: anchors in & \small $\sigma_s\!=\!16$ & \scriptsize $\sigma_s\!=\!4$ & \scriptsize difference \\
\includegraphics[width=0.235\linewidth]{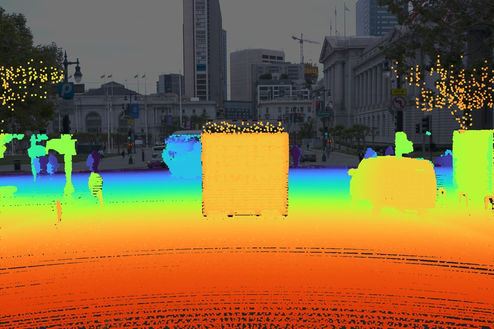} & \includegraphics[width=0.235\linewidth]{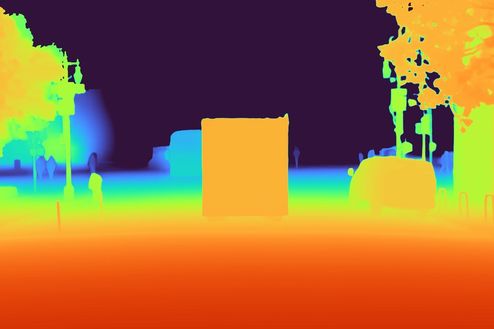} & \includegraphics[width=0.235\linewidth]{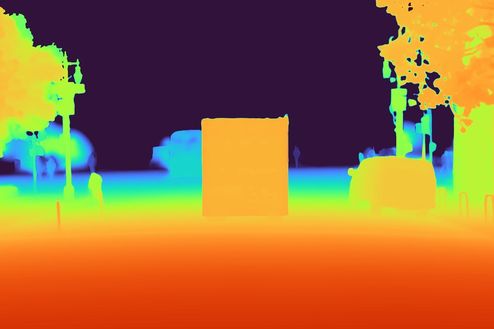} & \includegraphics[width=0.235\linewidth]{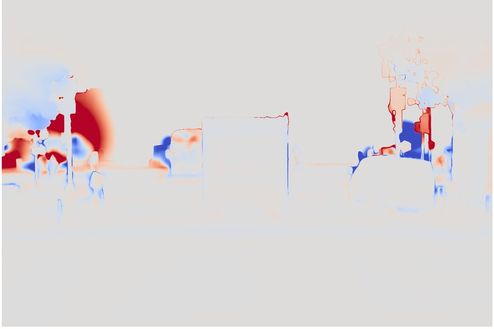} \\
\small Waymo: anchors in & \small $\sigma_s\!=\!16$ & \scriptsize $\sigma_s\!=\!4$ & \scriptsize difference \\
\includegraphics[width=0.235\linewidth]{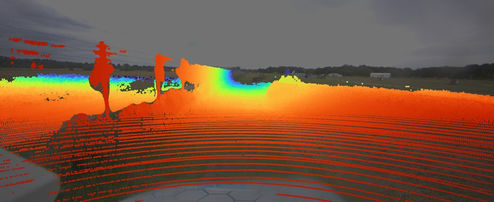} & \includegraphics[width=0.235\linewidth]{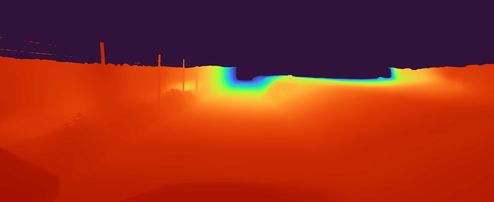} & \includegraphics[width=0.235\linewidth]{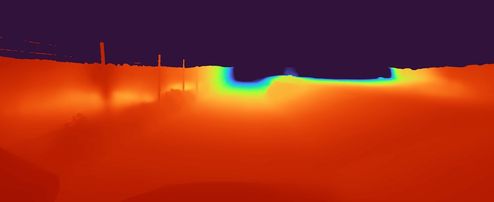} & \includegraphics[width=0.235\linewidth]{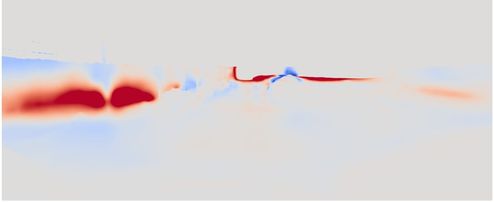} \\
\small GOOSE: anchors in & \small $\sigma_s\!=\!16$ & \scriptsize $\sigma_s\!=\!4$ & \scriptsize difference \\
\includegraphics[width=0.235\linewidth]{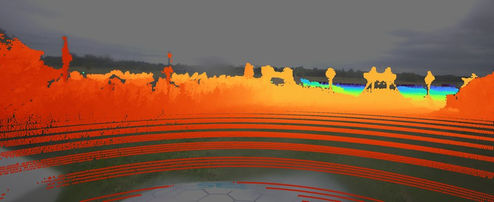} & \includegraphics[width=0.235\linewidth]{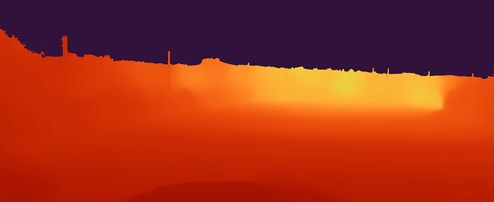} & \includegraphics[width=0.235\linewidth]{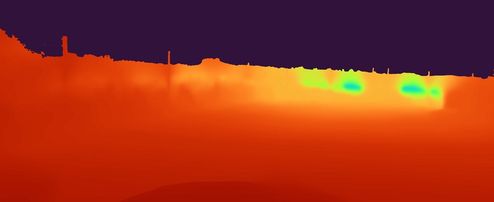} & \includegraphics[width=0.235\linewidth]{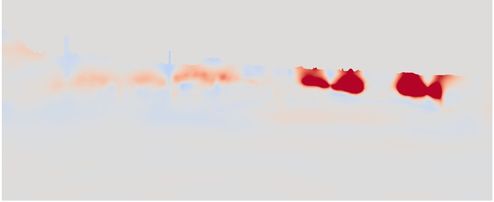} \\
\small GOOSE: anchors in & \small $\sigma_s\!=\!16$ & \scriptsize $\sigma_s\!=\!4$ & \scriptsize difference \\
\end{tabular}
\caption{Solver sensitivity, two frames per dataset: the anchors given to the solver, depth at the
default $\sigma_s\!=\!16$, at $\sigma_s\!=\!4$, and their signed difference (blue nearer, red farther,
grey unchanged, each panel on its own scale). At $\sigma_s\!=\!4$ the correction concentrates into blobs
at individual anchors instead of travelling along surfaces --- the $21$--$33\%$ RMSE it buys is paid for
in surface damage. NYU is omitted: its two settings differ by under a centimetre.}
\label{fig:apx-trade}
\end{figure*}

Figure~\ref{fig:apx-trade} is why we keep it anyway. At $\sigma_s\!=\!4$ the correction stops
travelling along road and facade surfaces and concentrates into small blobs at individual anchors: the
extra accuracy is bought by pulling the depth map locally towards each measurement, which is the surface
damage the method exists to avoid.

The structure metric cannot report it. Normal dispersion is the median angle between each surface normal
and its own $15\times15$ local mean over every valid pixel, so it describes the typical pixel, and anchor
blobs are a sparse minority that no median will follow. That robustness is deliberate --- the damage it
exists to catch in the learned baselines is global corrugation, for which the median is the right summary
--- but it leaves the metric blind to localised damage of our own, and a high quantile of the same angle
would be the natural way to expose it.

We take the coarser cell and report the point accuracy it gives up, since a depth map used for
reconstruction or 3D projection needs the surface rather than the last decimal of RMSE.

Table~\ref{tab:apx-parts} toggles each structural component of the solver alone, with runtime, so that a
component's contribution can be read against its cost. Solve time is dominated by the conjugate-gradient budget at half resolution, and moves only for the two
components that touch one of those: the iteration count ($27$ and $42$\,ms) and the solve resolution
($40$\,ms).

\begin{figure*}[!tbp]
\centering
\setlength{\tabcolsep}{0pt}
\begin{tabular}{@{}lc@{\hspace{1pt}}c@{\hspace{1pt}}c@{\hspace{1pt}}c@{\hspace{1pt}}c@{\hspace{1pt}}c@{\hspace{1pt}}c@{\hspace{1pt}}@{}}
 & \small RGB & \small ours & \small OMNI-DC & \small Prior-DA & \small BP-Net & \small NLSPN & \small DMD3C \\[1pt]
\rotatebox{90}{\scriptsize\ KITTI} & \includegraphics[width=0.1375\linewidth]{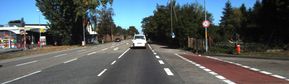} & \includegraphics[width=0.1375\linewidth]{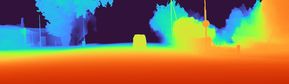} & \includegraphics[width=0.1375\linewidth]{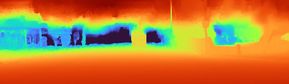} & \includegraphics[width=0.1375\linewidth]{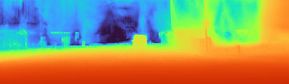} & \includegraphics[width=0.1375\linewidth]{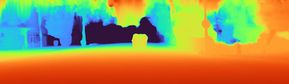} & \includegraphics[width=0.1375\linewidth]{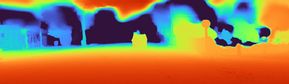} & \includegraphics[width=0.1375\linewidth]{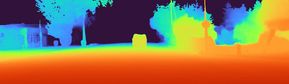} \\
\\[-4pt]
\rotatebox{90}{\scriptsize\ nuScenes} & \includegraphics[width=0.1375\linewidth]{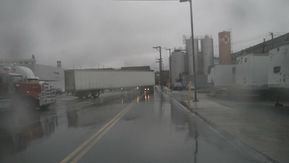} & \includegraphics[width=0.1375\linewidth]{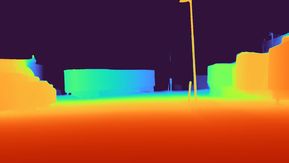} & \includegraphics[width=0.1375\linewidth]{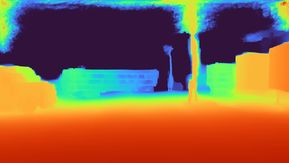} & \includegraphics[width=0.1375\linewidth]{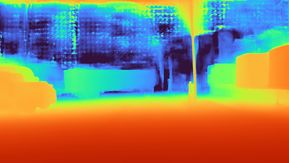} & \includegraphics[width=0.1375\linewidth]{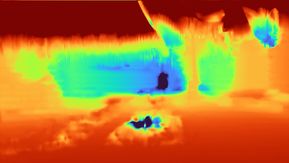} & \includegraphics[width=0.1375\linewidth]{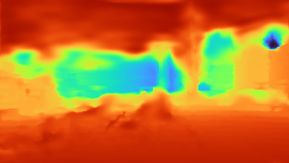} & \includegraphics[width=0.1375\linewidth]{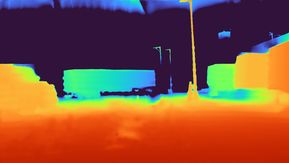} \\
\\[-4pt]
\rotatebox{90}{\scriptsize\ Waymo} & \includegraphics[width=0.1375\linewidth]{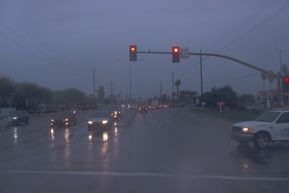} & \includegraphics[width=0.1375\linewidth]{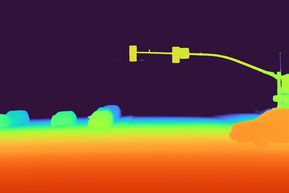} & \includegraphics[width=0.1375\linewidth]{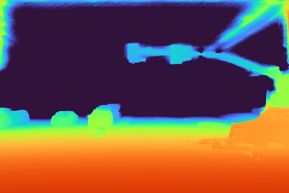} & \includegraphics[width=0.1375\linewidth]{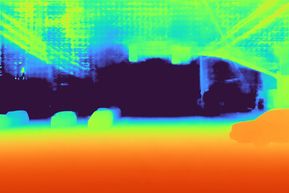} & \includegraphics[width=0.1375\linewidth]{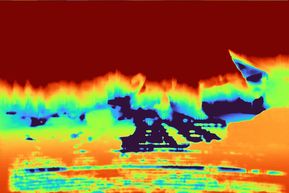} & \includegraphics[width=0.1375\linewidth]{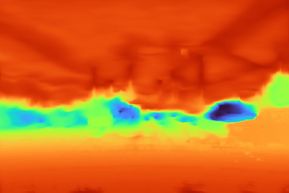} & \includegraphics[width=0.1375\linewidth]{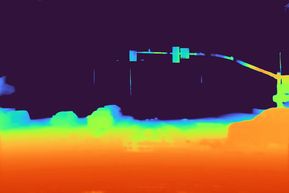} \\
\\[-4pt]
\rotatebox{90}{\scriptsize\ GOOSE} & \includegraphics[width=0.1375\linewidth]{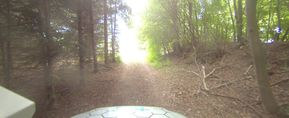} & \includegraphics[width=0.1375\linewidth]{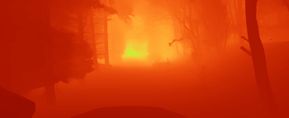} & \includegraphics[width=0.1375\linewidth]{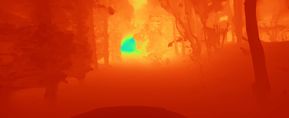} & \includegraphics[width=0.1375\linewidth]{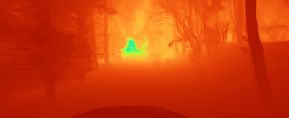} & \includegraphics[width=0.1375\linewidth]{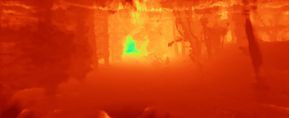} & \includegraphics[width=0.1375\linewidth]{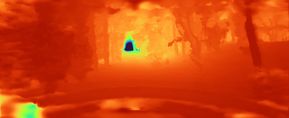} & \includegraphics[width=0.1375\linewidth]{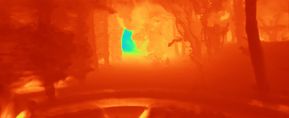} \\
\\[-4pt]
\rotatebox{90}{\scriptsize\ NYU} & \includegraphics[width=0.1375\linewidth]{appendix/images/v2_nyu_rgb.jpg} & \includegraphics[width=0.1375\linewidth]{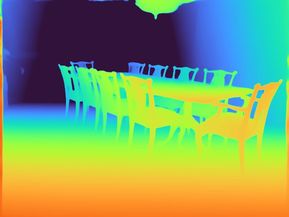} & \includegraphics[width=0.1375\linewidth]{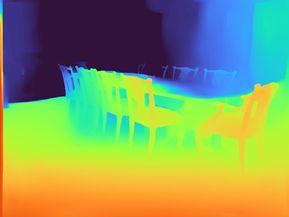} & \includegraphics[width=0.1375\linewidth]{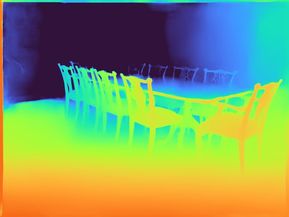} & \includegraphics[width=0.1375\linewidth]{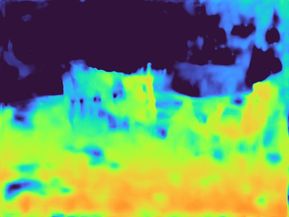} & \includegraphics[width=0.1375\linewidth]{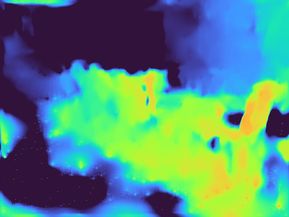} & \includegraphics[width=0.1375\linewidth]{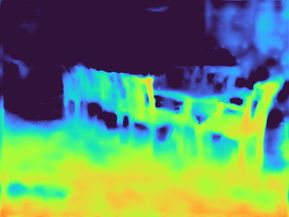} \\
\\[-4pt]
\end{tabular}
\caption{Predicted depth for every evaluated method on one frame of each dataset, alongside the input
image. Frames are drawn at random from each dataset, except GOOSE, where a forest track was chosen over the
open field the draw returned. Our sky and other pixels where the
monocular prior is undefined are drawn at the depth cap, since we emit no prediction there. The supervised
networks hold up on the driving scenes that resemble their training data and come apart away from them,
most visibly indoors. On the driving rows every method returns a plausible map --- and on those same
frames their normal dispersion spans $1.7$ to $56$ degrees (Figure~\ref{fig:apx-normals}). What separates
these methods is not visible in this representation.}
\label{fig:apx-depths}
\end{figure*}

\begin{figure*}[t]
\centering
\setlength{\tabcolsep}{0pt}
\begin{tabular}{@{}lc@{\hspace{1pt}}c@{\hspace{1pt}}c@{\hspace{1pt}}c@{\hspace{1pt}}c@{\hspace{1pt}}c@{\hspace{1pt}}c@{\hspace{1pt}}@{}}
 & \small anchors in & \small ours & \small OMNI-DC & \small Prior-DA & \small BP-Net & \small NLSPN & \small DMD3C \\[1pt]
\rotatebox{90}{\scriptsize\ KITTI} & \includegraphics[width=0.1375\linewidth]{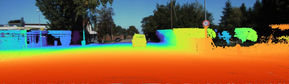} & \includegraphics[width=0.1375\linewidth]{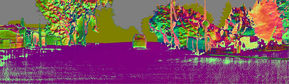} & \includegraphics[width=0.1375\linewidth]{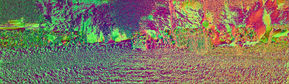} & \includegraphics[width=0.1375\linewidth]{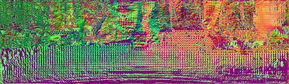} & \includegraphics[width=0.1375\linewidth]{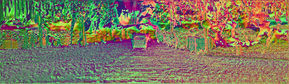} & \includegraphics[width=0.1375\linewidth]{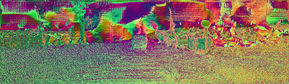} & \includegraphics[width=0.1375\linewidth]{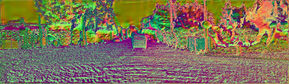} \\
 &  & \scriptsize 1.02/5.7 & \scriptsize 1.46/48.4 & \scriptsize 2.77/40.7 & \scriptsize 0.62/48.0 & \scriptsize 2.10/39.7 & \scriptsize 0.57/38.3 \\[1.5pt]
\rotatebox{90}{\scriptsize\ nuScenes} & \includegraphics[width=0.1375\linewidth]{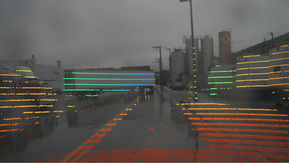} & \includegraphics[width=0.1375\linewidth]{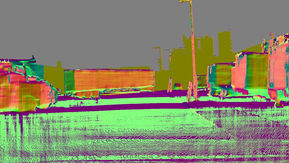} & \includegraphics[width=0.1375\linewidth]{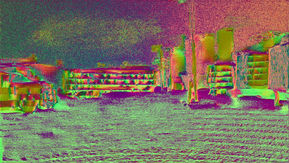} & \includegraphics[width=0.1375\linewidth]{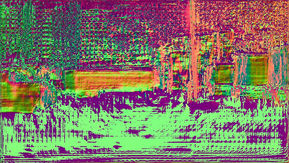} & \includegraphics[width=0.1375\linewidth]{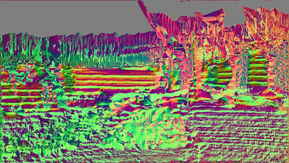} & \includegraphics[width=0.1375\linewidth]{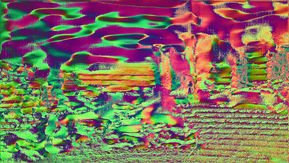} & \includegraphics[width=0.1375\linewidth]{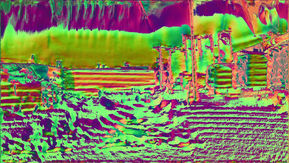} \\
 &  & \scriptsize 1.06/5.3 & \scriptsize 0.16/29.6 & \scriptsize 0.92/13.9 & \scriptsize 0.83/29.9 & \scriptsize 0.45/18.5 & \scriptsize 0.36/13.7 \\[1.5pt]
\rotatebox{90}{\scriptsize\ Waymo} & \includegraphics[width=0.1375\linewidth]{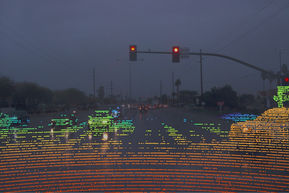} & \includegraphics[width=0.1375\linewidth]{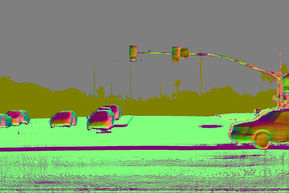} & \includegraphics[width=0.1375\linewidth]{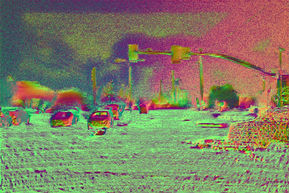} & \includegraphics[width=0.1375\linewidth]{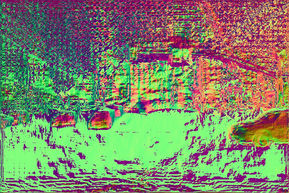} & \includegraphics[width=0.1375\linewidth]{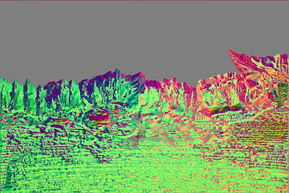} & \includegraphics[width=0.1375\linewidth]{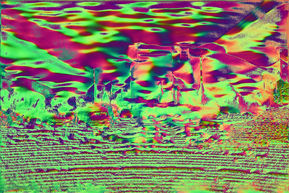} & \includegraphics[width=0.1375\linewidth]{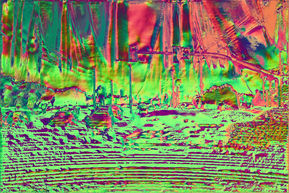} \\
 &  & \scriptsize 1.15/1.7 & \scriptsize 0.50/30.2 & \scriptsize 1.22/15.8 & \scriptsize 8.84/56.1 & \scriptsize 0.46/16.9 & \scriptsize 0.50/19.9 \\[1.5pt]
\rotatebox{90}{\scriptsize\ GOOSE} & \includegraphics[width=0.1375\linewidth]{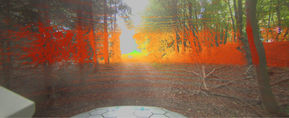} & \includegraphics[width=0.1375\linewidth]{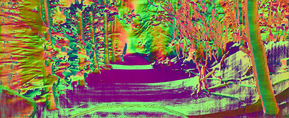} & \includegraphics[width=0.1375\linewidth]{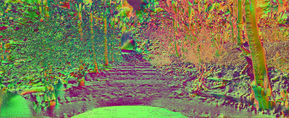} & \includegraphics[width=0.1375\linewidth]{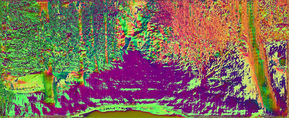} & \includegraphics[width=0.1375\linewidth]{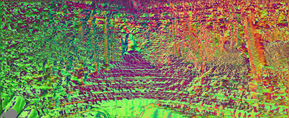} & \includegraphics[width=0.1375\linewidth]{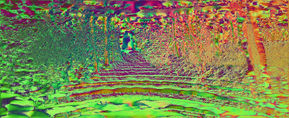} & \includegraphics[width=0.1375\linewidth]{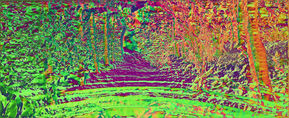} \\
 &  & \scriptsize 0.95/12.1 & \scriptsize 0.47/33.4 & \scriptsize 0.62/16.8 & \scriptsize 0.55/30.1 & \scriptsize 0.54/29.6 & \scriptsize 0.55/22.6 \\[1.5pt]
\rotatebox{90}{\scriptsize\ NYU} & \includegraphics[width=0.1375\linewidth]{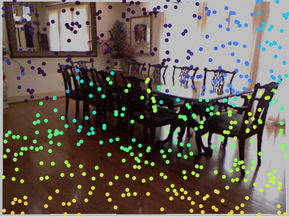} & \includegraphics[width=0.1375\linewidth]{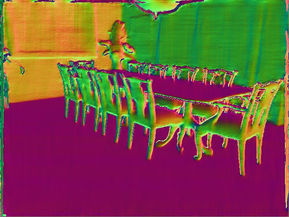} & \includegraphics[width=0.1375\linewidth]{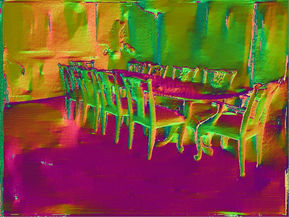} & \includegraphics[width=0.1375\linewidth]{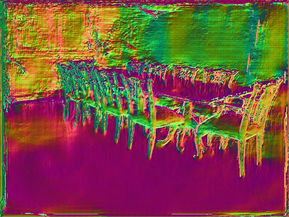} & \includegraphics[width=0.1375\linewidth]{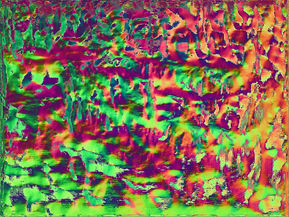} & \includegraphics[width=0.1375\linewidth]{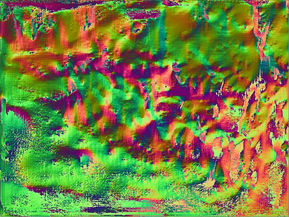} & \includegraphics[width=0.1375\linewidth]{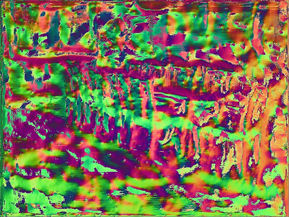} \\
 &  & \scriptsize 0.34/6.3 & \scriptsize 0.20/8.1 & \scriptsize 0.27/11.8 & \scriptsize 0.66/23.8 & \scriptsize 1.27/17.6 & \scriptsize 0.79/21.5 \\[1.5pt]
\end{tabular}
\caption{Surface normals of the same predictions, with each panel's RMSE\,(m) and normal dispersion\,(deg)
beneath it. The completion networks imprint the sensor into the surface: the ruling visible across the road
is the LiDAR sweep pattern of the anchor column reproduced as geometry. The point metric does not register
it --- DMD3C and BP-Net are both more accurate than we are on the KITTI frame while carrying seven to eight times
the normal error, because reproducing a measurement exactly is rewarded by RMSE whether or not the surface
between measurements survives. We estimate a smooth metric correction over a prior we do not modify, so the
anchor pattern cannot enter the geometry in the first place.}
\label{fig:apx-normals}
\end{figure*}

\begin{figure*}[!tbp]
\centering
\begin{tabular}{@{}c@{\hspace{1.5pt}}c@{\hspace{1.5pt}}c@{\hspace{1.5pt}}c@{\hspace{1.5pt}}c@{}}
\includegraphics[width=0.195\linewidth]{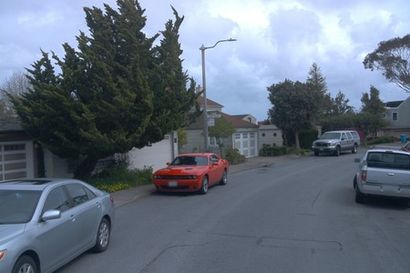} & \includegraphics[width=0.195\linewidth]{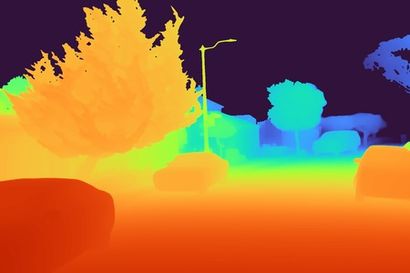} & \includegraphics[width=0.195\linewidth]{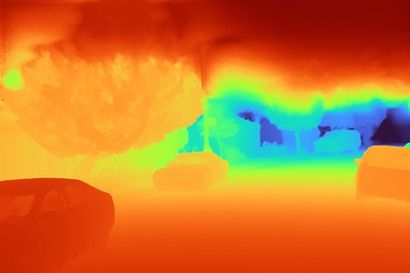} & \includegraphics[width=0.195\linewidth]{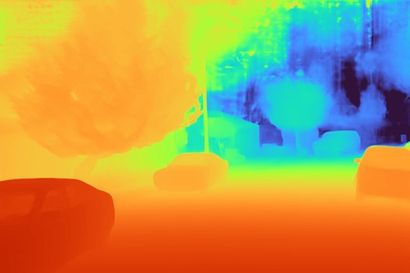} & \includegraphics[width=0.195\linewidth]{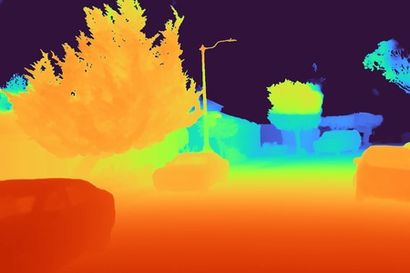} \\
\small Waymo: RGB & \small ours & \small OMNI-DC & \small Prior-DA & \small DMD3C \\[1pt]
\includegraphics[width=0.195\linewidth]{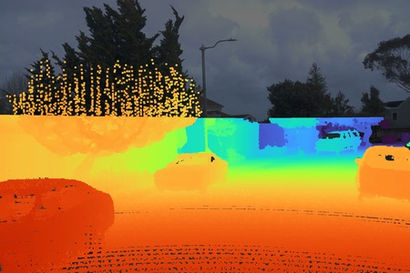} & \includegraphics[width=0.195\linewidth]{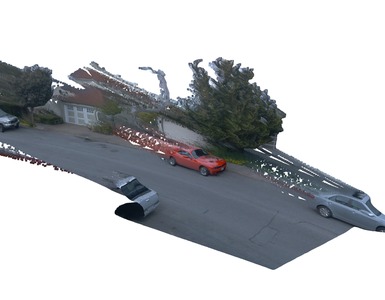} & \includegraphics[width=0.195\linewidth]{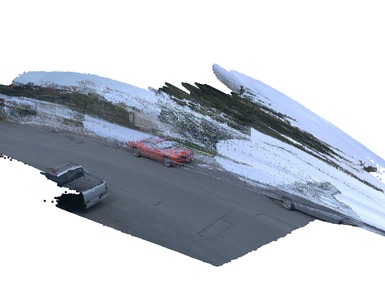} & \includegraphics[width=0.195\linewidth]{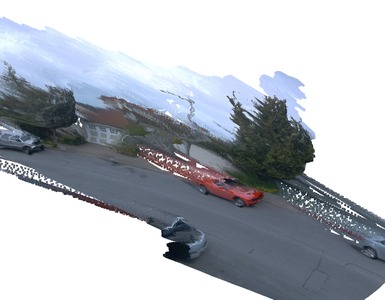} & \includegraphics[width=0.195\linewidth]{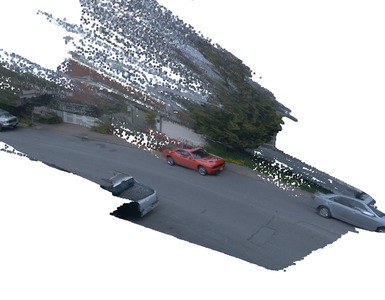} \\
\small anchors on RGB & \small ours & \small OMNI-DC & \small Prior-DA & \small DMD3C \\[5pt]
\includegraphics[width=0.195\linewidth]{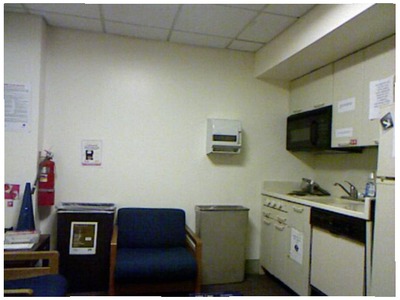} & \includegraphics[width=0.195\linewidth]{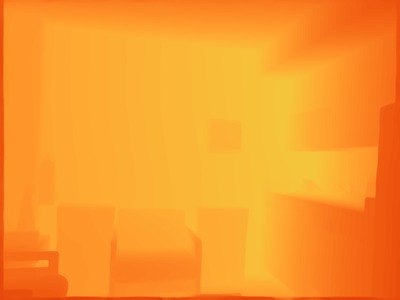} & \includegraphics[width=0.195\linewidth]{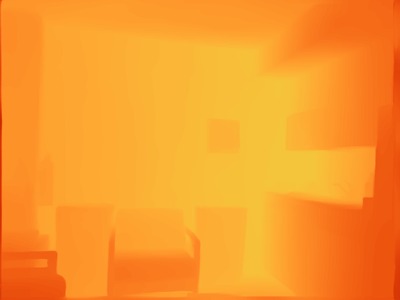} & \includegraphics[width=0.195\linewidth]{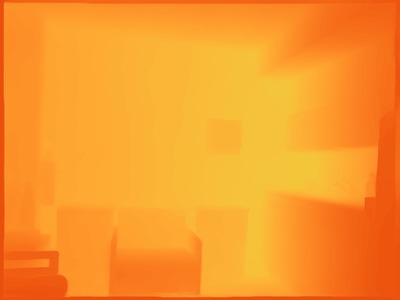} & \includegraphics[width=0.195\linewidth]{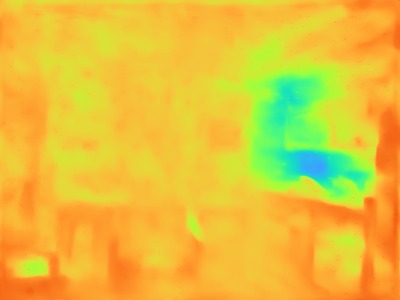} \\
\small NYU: RGB & \small ours & \small OMNI-DC & \small Prior-DA & \small DMD3C \\[1pt]
\includegraphics[width=0.195\linewidth]{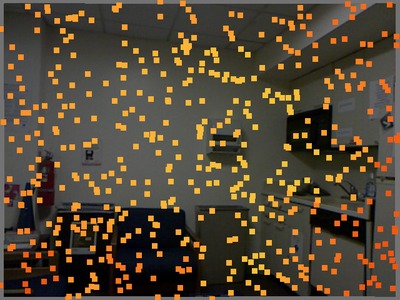} & \includegraphics[width=0.195\linewidth]{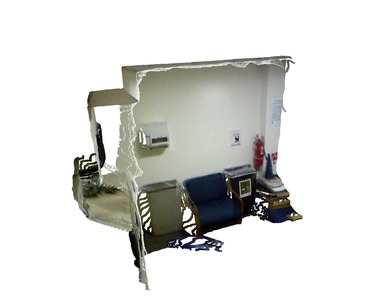} & \includegraphics[width=0.195\linewidth]{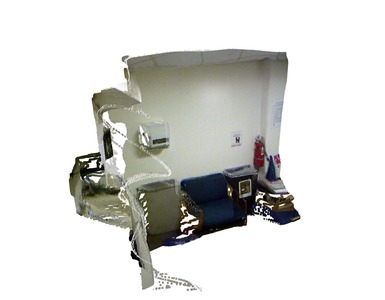} & \includegraphics[width=0.195\linewidth]{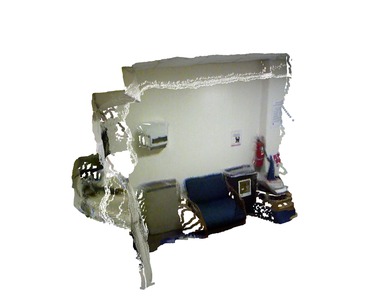} & \includegraphics[width=0.195\linewidth]{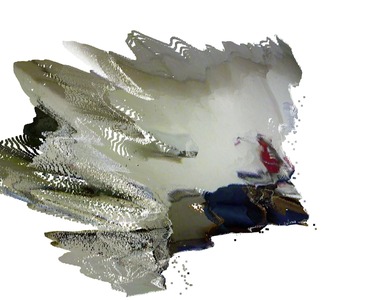} \\
\small anchors on RGB & \small ours & \small OMNI-DC & \small Prior-DA & \small DMD3C \\[5pt]
\end{tabular}
\caption{The same depth maps, turned back into 3D. For each scene the upper row is the input image
and each method's depth map; the lower row is the anchors the solver was given, drawn over the image, and each depth map re-projected
into a coloured point cloud, all seen from a direction the camera never occupied ($35^\circ$ azimuth,
$18^\circ$ elevation). From the original viewpoint all four look alike by construction, since each is
consistent with the same image. The depth maps are indeed hard to separate --- but only ours re-projects
to continuous surfaces. On Waymo the baselines smear the buildings and trees behind the road into sheets of stray
points that hang over the scene; on NYU, where OMNI-DC and Prior-DA are
metrically \emph{ahead} of us, the room still resolves for all three and only DMD3C disintegrates. This is
what the normal-dispersion column measures, and what a downstream consumer of the depth map receives.}
\label{fig:apx-cloud}
\end{figure*}

Figures~\ref{fig:apx-depths}, \ref{fig:apx-normals} and \ref{fig:apx-cloud} close the
appendix with every method on the same frames, seen three ways: as predicted depth, as the
surface normals of those same predictions, and re-projected to 3D. The depth maps are the
representation in which the methods are hardest to tell apart, and the clouds the one in which
they are easiest.

\end{document}